\PassOptionsToPackage{table}{xcolor}
\documentclass[manuscript,nonacm]{acmart}

\usepackage{booktabs}
\usepackage{tabularx}
\usepackage{array}
\usepackage{multirow}
\usepackage[table]{xcolor}

\newcolumntype{Y}[1]{>{\hsize=#1\hsize\linewidth=\hsize\raggedright\arraybackslash}X}

\acmJournal{IMWUT}

\AtBeginDocument{%
  }

\begin{document}
% \begin{CJK*}{UTF8}{gbsn}
\title{When Optimization Meets Human Intent: Human-Centered Diffusion Model Optimization}

\author{Renye Yan}
\affiliation{%
  \institution{Peking University}
  \country{China}
}

\author{Jikang Cheng}
\affiliation{%
  \institution{Peking University}
  \country{China}
}

% \author{Yi Gao}
% \affiliation{%
%   \institution{Peking University}
%   \country{China}
% }

% \author{Yanzeng Li}
% \affiliation{%
%   \institution{Peking University}
%   % \city{Beijing}
%   \country{China}
% }

% % Ling Liang and Yimao Cai are co-corresponding authors.

% \author{Ling Liang}
% % \authornote{Ling Liang is the corresponding authors.}
% \affiliation{%
%   \institution{Peking University}
%   % \city{Beijing}
%   \country{China}
% }

% \author{Jialun Zhong}
% \authornote{Jialun Zhong is the corresponding authors.}
% \affiliation{%
%   \institution{Peking University}
%   % \city{Beijing}
%   \country{China}
% }

\begin{abstract}

Diffusion models are evolving from general tools into personalized systems that adapt to user needs through fine-tuning and guidance. However, diffusion optimization mainly targets measurable objectives such as aesthetics, rarely examining whether they align with human intent. Human needs are multidimensional, contextual, and dynamic, whereas existing methods approximate them with fixed metrics. Because these metrics capture human needs only partially and may amplify deviations from true intent during optimization, higher scores need not yield results better matching human expectations. To understand how human needs enter and influence optimization, we systematically analyze how human intent is expressed, translated into optimizable objectives, applied to model behavior or generation, and evaluated through outputs. We further discuss trade-offs affecting human-intent alignment in personalized image generation and gaps between automated metrics and real user judgments. This human-centered synthesis reorganizes diffusion optimization and provides a systematic reference for personalized generation systems designed for human preferences.
\end{abstract}

\begin{CCSXML}
<ccs2012>
   <concept>
       <concept_id>10010147.10010178.10010224</concept_id>
       <concept_desc>Computing methodologies~Computer vision</concept_desc>
       <concept_significance>500</concept_significance>
       </concept>
   <concept>
       <concept_id>10003120.10003121.10003124.10011751</concept_id>
       <concept_desc>Human-centered computing~Collaborative interaction</concept_desc>
       <concept_significance>500</concept_significance>
       </concept>
 </ccs2012>
\end{CCSXML}

\ccsdesc[500]{Computing methodologies~Computer vision}
\ccsdesc[500]{Human-centered computing~Collaborative interaction}

\keywords{Human–AI Co-Creation,  Human Intent, Human-Centered AI, Generative Model Optimization}

% ————————————————————————————————————————————————————————————————————
\begin{teaserfigure}
  \includegraphics[width=\textwidth]{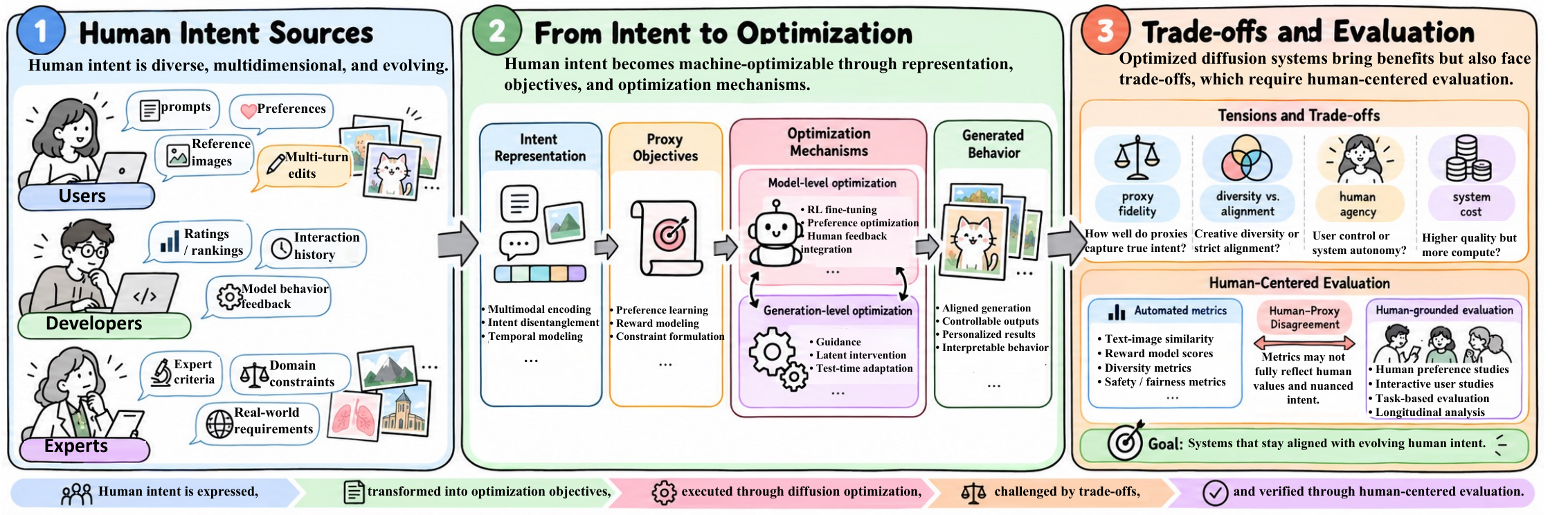}
  \caption{The framework illustrates how human intent originates from users and creators, researchers and developers, and domain experts; is represented and translated into machine-optimizable objectives; and subsequently influences diffusion model behavior through model-level and generation-level optimization. It further highlights the key trade-offs arising during optimization, including proxy fidelity, diversity–alignment balance, human agency, and system cost. Finally, it summarizes complementary evaluation practices, ranging from automated metrics to human-grounded evaluations, for assessing whether optimization outcomes genuinely align with evolving human intent..}
  % \Description{Enjoying the baseball game from the third-base
  % seats. Ichiro Suzuki preparing to bat.}
  \label{fig:teaser}
\end{teaserfigure}
% ————————————————————————————————————————————————————————————————————


\maketitle

\section{Introduction}
\label{sec:intro}

% Research on diffusion models~\cite{huang2024plantography,chen2024chatscratch,xiao2024typedance,choi2024creativeconnect,son2024genquery,dhawka2024better,rajcic2024towards,zhang2024partiality,lee2024impact,kamali2025characterizing,porquet2025copying} is shifting from `how to generate images' to `how to better meet specific human preferences'~\cite{wu2026preference,liu2026alignment}. In earlier text-to-image paradigms, users expressed intent mainly through prompts, and models generated images accordingly. Recently, fine-tuning~\cite{black2024training} and training-free guidance~\cite{yan2026swiftexplorer} have changed this interaction: generative models no longer merely respond to text prompts, but can continuously adapt their behavior using human preferences and feedback to optimize toward specific goals~\cite{black2023training,wallace2024diffusion}.
% Although above personalized generation is becoming increasingly common, models cannot directly turn user needs into personalized images. From a user’s request to an image aligned with their preferences, the process involves at least three related but distinct stages.

Research on diffusion models~\cite{huang2024plantography,chen2024chatscratch,xiao2024typedance,choi2024creativeconnect,son2024genquery,dhawka2024better,rajcic2024towards,zhang2024partiality,lee2024impact,kamali2025characterizing,porquet2025copying} is shifting from ``how to generate images'' to ``how to better meet specific human preferences''~\cite{wu2026preference,liu2026alignment}. In earlier text-to-image paradigms, users expressed intent mainly through prompts, and models generated images accordingly. Recently, fine-tuning~\cite{black2024training} and training-free guidance~\cite{yan2026swiftexplorer} have changed this interaction: generative models no longer merely respond to text prompts, but can adapt their behavior based on human preferences and feedback to optimize toward specific goals~\cite{black2023training,wallace2024diffusion}.

Despite these advances, optimizing generative models toward human preferences does not necessarily ensure that their outputs faithfully reflect human intent~\cite{dang2025personalized}. Human needs are often multidimensional, subjective, and context-dependent, making it difficult to directly translate them into computational objectives and verify whether the resulting images truly satisfy user expectations~\cite{zhang2024learning}. Consequently, bridging the gap between human intent and personalized image generation involves three interconnected challenges.

\textbf{(1) \textit{User intent must be expressed and translated into computable objectives}.}
User intent must first be externalized through text, choices, ratings, or other forms of feedback, and subsequently translated into machine-optimizable objectives~\cite{shen2026interaction,xu2023imagereward}. Thus, what optimization algorithms actually process is not human intent itself, but a proxy representation constructed from these interaction signals~\cite{zeng2024intenttuner,subramonyam2024bridging}. Different forms of expression capture different aspects of user needs, and the same subjective need may be mapped to different computational objectives, potentially resulting in information loss before intent enters the optimization process~\cite{mahdavi2024ai,fan2024contextcam}.

\textbf{(2) \textit{Different user needs must be coordinated as optimization objectives}.} Personalized generation often involves multiple dimensions, such as aesthetic quality, semantic consistency, realism, and diversity~\cite{zhang2024learning,fan2024contextcam}. These dimensions may naturally compete with each other, and their importance can change across tasks and generated results~\cite{zhou2024understanding,maier2026partnering}. However, existing methods usually combine multiple evaluation objectives in a static way, compressing users’ multidimensional and dynamic needs into optimization signals with fixed weights~\cite{lee2025calibrated,lee2026flow}. Such preset objective relations are hard to adjust with changing user preferences~\cite{cheng2026diffusion}. As a result, the model may keep reinforcing some objectives while sacrificing others that are more important in the current context, causing the optimization direction to deviate from users’ actual preferences~\cite{jena2024elucidating,hong2026understanding,amershi2019guidelines}. 

\textbf{(3) \textit{Optimization outcomes must be validated against actual human preferences}.}
Improving predefined optimization objectives does not necessarily mean that generated results better satisfy human needs or align with user intent~\cite{wadinambiarachchi2024effects,jayasumana2024rethinking}. Existing studies often evaluate generated outputs using reward scores, text-image alignment, aesthetic metrics, or other automated evaluators, which may share similar assumptions with the objectives used during optimization~\cite{wiles2025revisiting,li2024evaluating}. Consequently, even when both optimization targets and evaluation metrics improve, it remains unclear whether these improvements correspond to what users actually perceive, prefer, or expect from the generated results~\cite{ma2025hpsv3,naqvi2025catalyst}. From a human-centered perspective, evaluating personalized generation requires moving beyond automated metrics to consider human judgments, individual preferences, and the specific contexts in which users interact with generative systems~\cite{wang2024roomdreaming,wang2024promptcharm}. For highly subjective and context-dependent needs, the key is therefore not merely to confirm improvements in proxy metrics, but to examine whether these gains translate into greater satisfaction of human preferences and better alignment with users' actual intentions~\cite{cao2025compositional}.

Together, these three challenges highlight that human intent is not directly mapped to the final output of a generative system; instead, it undergoes a continuous transformation process. Human intent must be represented as machine-optimizable objectives, which are then coordinated and translated into changes in model behavior or generation trajectories~\cite{fu2025chats,shen2025ideationweb}. Throughout this process, incomplete proxy representations, competing objectives, and optimization mechanisms may introduce tensions and unintended outcomes, potentially weakening alignment with human preferences. Ultimately, whether optimization truly succeeds must be assessed through human-centered validation rather than improvements in automated metrics alone~\cite{huh2025videodiff,trinh2025picture}. These interconnected challenges suggest that understanding human-centered diffusion model optimization requires examining not only how optimization is performed, but also how human intent is represented, transformed, potentially distorted, and ultimately evaluated. Therefore, rather than organizing existing studies by algorithm type~\cite{jiang2024survey,cao2025controllable}. This paper re-examines diffusion model optimization through four interconnected dimensions: intent representation, optimization mechanisms, failure modes and trade-offs, and human-centered validation. Based on this perspective, we address the following four research questions:

\begin{itemize}
    \item RQ1: How is human intent represented and operationalized in diffusion model optimization?

    \item RQ2: How do existing optimization paradigms translate human intent into changes in model behavior or generation trajectories?

    \item RQ3: What tensions and failure modes emerge when optimizing diffusion models toward human-defined or human-proxied objectives?

    \item RQ4: How is alignment with human intent validated through automated and human-centered evaluations, and what limitations remain in current evaluation practices?
\end{itemize}

\begin{figure}
    \centering
    \includegraphics[width=1.0\linewidth]{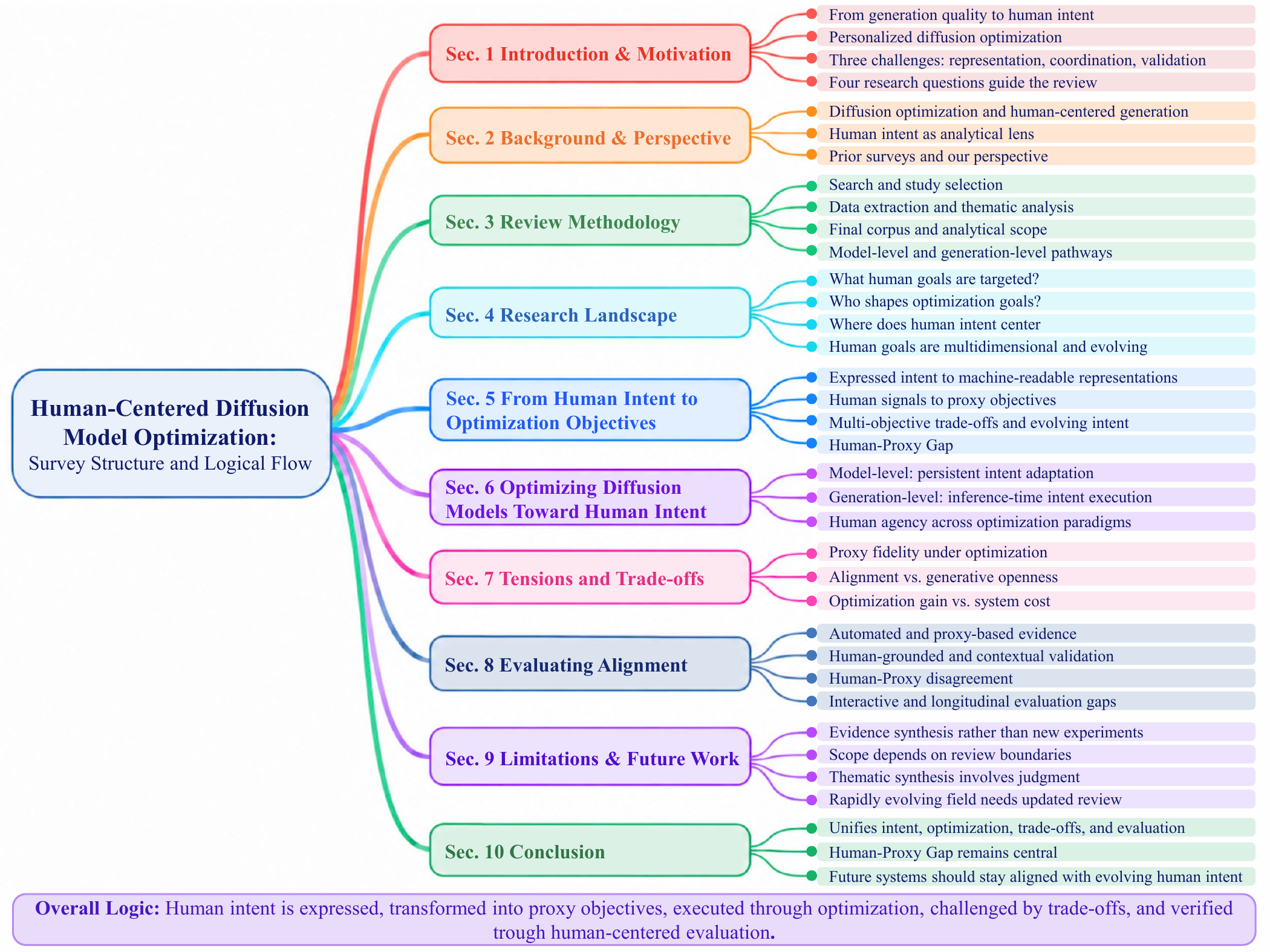}
    \caption{Overview of the Survey and Research Roadmap}
    \label{paper_structure}
\end{figure}

To address these research questions, we systematically review existing studies on diffusion model optimization through a human-centered analytical framework encompassing four interconnected dimensions: intent representation, optimization mechanisms, failure modes and trade-offs, and human-centered validation. Specifically, we investigate how human intent is expressed and translated into machine-optimizable objectives (RQ1), how different optimization paradigms translate these objectives into changes in model behavior or generation trajectories (RQ2), what tensions and failure modes emerge during optimization (RQ3), and how existing evaluation practices assess whether optimization outcomes genuinely align with human preferences and user needs (RQ4). By connecting these dimensions, our review provides a unified perspective on how human intent is represented, operationalized, executed, and validated throughout diffusion model optimization. The main contributions of this paper are as follows:
\begin{itemize}

\item We systematically examine how human intent is expressed, represented, and translated into machine-optimizable objectives, revealing potential information loss and mismatches between user needs and computational proxies.

\item We organize reinforcement-learning fine-tuning and inference-time optimization into model-level and generation-level paradigms, systematically comparing how they translate human-centered objectives into changes in model behavior or generation trajectories.

\item We identify key trade-offs and failure modes in human-centered diffusion model optimization, examining how proxy-objective mismatch, competing objectives, reduced diversity, and optimization costs may undermine alignment with human intent.

\item We systematically examine automated and human-centered evaluation practices, highlighting discrepancies between proxy-based metrics and actual human judgments, and identifying limitations in validating whether optimization outcomes genuinely satisfy user needs.
\end{itemize}
The overall structure of this survey and the logical relationships among its sections are summarized in \textbf{Fig}.~\ref{paper_structure}.

% Required packages
% \usepackage{booktabs}
% \usepackage{tabularx}
% \usepackage{array}

\begin{table*}[t]
\centering
\caption{Proposed Taxonomy Tables Across the Survey Chapters}
\label{tab:proposed_taxonomy_tables}

\small
\renewcommand{\arraystretch}{1.25}
\setlength{\tabcolsep}{7pt}

\begin{tabularx}{\textwidth}{
    >{\raggedright\arraybackslash}p{0.12\textwidth}
    >{\raggedright\arraybackslash}p{0.43\textwidth}
    >{\raggedright\arraybackslash}X
}

\toprule
\textbf{Section} &
\textbf{Proposed Table Title} &
\textbf{Main Classification Dimensions} \\
\midrule

Sec.~\ref{Landscape} &
Stakeholders and Their Roles in Shaping Optimization Goals &
End users, researchers, domain experts, and computational proxies. \\

\midrule

Sec.~\ref{Human_Intent_to_Optimization} &
Taxonomy of Human Intent Operationalization &
Transformation of human intent into constraints, reward signals,
preference models, and multi-objective optimization signals. \\

\midrule

Sec.~\ref{Optimizing_Diffusion_Models} &
Taxonomy of Human-Centered Diffusion Optimization Methods &
Reinforcement learning fine-tuning, direct preference optimization
(DPO), training-free guidance, and latent/trajectory optimization. \\

\midrule

Sec.~\ref{section7} &
Taxonomy of Optimization Trade-offs and Failure Modes &
Reward hacking, diversity loss, human agency,
and computational and interaction costs. \\

\midrule

Sec.~\ref{section8} &
Taxonomy of Evaluation Practices for Human-Intent Alignment &
Automated metrics, preference-based evaluation,
human judgments, expert evaluation,
and interactive and longitudinal evaluation. \\

\midrule

Sec.~\ref{section9} &
Research Gaps and Future Directions &
Adaptive intent modeling, dynamic objective coordination,
user control, and interactive evaluation. \\

\bottomrule
\end{tabularx}

\end{table*}

\section{Related Work}

\subsection{Human Intent in Diffusion Model Optimization}

Diffusion models~\cite{smith2026noise,zhao2026less,grace2026entropic,li2026preference,simonen2026exploration,wang2023reprompt,moore2023failurenotes,yan2026timing} generate images through iterative denoising, allowing external signals to influence intermediate states and offering more control than one-shot conditioning. Early classifier and classifier-free guidance show that changing sampling directions can shape outputs and trade off quality, alignment, and diversity.
Recently, such control has moved toward preference-driven personalized generation. Some methods update model parameters to change long-term behavior~\cite{zhao2022alphaholdem,gan2024transductive,yan2026past,yan2026explore}, while others keep parameters fixed and adjust latents or denoising trajectories at inference time. We call them \textit{model-level optimization} and \textit{generation-level optimization}.
From an HCI view, the key is not only what reward is used, but how human needs enter the system~\cite{yan2026explicitintentdiscoveringshaping}. In generative AI, user goals are often clarified through repeated generation, observation, and revision~\cite{amershi2019guidelines,brade2023promptify,wang2024promptcharm}. Prompts are only one expression form and are limited for fine-grained or contextual needs; choices, clicks, and context can further convey intent~\cite{shen2026interaction,fan2024contextcam}. Thus, human intent refers to the user’s overall goal for both the desired output and the generation process, while interaction signals are its external representations~\cite{brade2023promptify}.
We define \textit{human-centered diffusion model optimization} as obtaining goals related to human needs and using them to actively change model behavior or generation trajectories. These needs may come directly from users or through proxy models~\cite{xu2023imagereward,black2023training,prabhudesai2023aligning,bansal2024universal,yanlitexplorer}.

\subsection{Human-Centered Diffusion Model Optimization}

Diffusion model optimization has evolved from improving generation capabilities and controllability toward aligning generated outputs with human preferences and intentions. Early research primarily focused on the fundamental principles and generation capabilities of text-to-image diffusion models~\cite{zhang2023text}. Subsequent studies extended these capabilities through controllable generation, introducing different mechanisms for incorporating specific conditions, multiple constraints, and general control signals into the generation process~\cite{jiang2024survey,cao2025controllable}. While these advances have improved the controllability of diffusion models, they primarily address how generation can be technically controlled rather than whether the resulting outputs genuinely satisfy human needs.

More recently, diffusion model optimization has increasingly incorporated human preferences and feedback into the generation process. Research has explored preference modeling, reinforcement learning from human feedback (RLHF), direct preference optimization (DPO), and test-time alignment to steer generative models toward human-defined objectives~\cite{liu2026alignment}. Related studies have examined preference alignment in image generation and editing, highlighting the roles of reward modeling, RLHF, and DPO in adapting generation behavior to human preferences~\cite{wu2026preference}. These developments suggest a growing shift from optimizing predefined technical objectives toward incorporating human preferences into model optimization. However, human preferences are often represented through computational proxies, raising questions about whether improvements in these objectives genuinely reflect users' underlying intentions.

From a human-centered computing perspective, effective optimization requires not only improving predefined objectives but also understanding how human intent is expressed, interpreted, and incorporated into generative systems. Prior HCI research emphasizes that user goals are not necessarily fixed or fully specified in advance; instead, they may be progressively clarified and revised through interaction, feedback, and contextual information~\cite{amershi2019guidelines,shen2026interaction,fan2024contextcam}. Accordingly, human-centered diffusion model optimization should be understood as a process that connects human intent with computational objectives, uses these objectives to influence model behavior or generation trajectories, and evaluates whether the resulting outputs genuinely satisfy user needs. This perspective moves beyond classifying optimization techniques solely by their technical mechanisms and instead emphasizes how human intent is represented, operationalized, and maintained throughout the optimization process.

\begin{table*}[t]
\centering
\caption{Comparison of representative surveys on diffusion models, controllable generation, and preference alignment with our human-centered perspective.}
\label{tab:survey_comparison}
\small
\renewcommand{\arraystretch}{1.25}
\setlength{\tabcolsep}{5pt}

\begin{tabularx}{\textwidth}{
    >{\raggedright\arraybackslash}p{0.16\textwidth}
    >{\raggedright\arraybackslash}p{0.19\textwidth}
    >{\raggedright\arraybackslash}X
    >{\raggedright\arraybackslash}X
}
\toprule
\textbf{Survey} &
\textbf{Primary Perspective} &
\textbf{Organizing Framework} &
\textbf{Central Research Focus} \\
\midrule

Zhang et al.~\cite{zhang2023text} &
Text-to-image diffusion &
Generative foundations, methods, applications, and challenges &
How diffusion models generate images from text \\
\midrule

Jiang et al.~\cite{jiang2024survey} &
Controllable diffusion generation &
Control formulations, mechanisms, and evaluation &
How diffusion generation can be guided by control signals \\
\midrule

Cao et al.~\cite{cao2025controllable} &
Controllable text-to-image generation &
Specific, multiple, and universal control conditions &
How different conditions are incorporated into generation \\
\midrule

Liu et al.~\cite{liu2026alignment} &
Human alignment of diffusion models &
RLHF, DPO, test-time alignment, benchmarks, and evaluation &
How diffusion models are aligned with human preferences \\
\midrule

Wu et al.~\cite{wu2026preference} &
Preference alignment in image generation and editing &
Reward modeling, RLHF, DPO, applications, and challenges &
How preference-alignment techniques improve generation and editing \\
\midrule

\textbf{This Review} &
\textbf{Human-intent transformation} &
\textbf{Intent representation $\rightarrow$ Proxy objectives
$\rightarrow$ Optimization $\rightarrow$ Trade-offs
$\rightarrow$ Human-centered validation} &
\textbf{How human intent is represented, transformed,
potentially distorted, and validated throughout optimization} \\

\bottomrule
\end{tabularx}
\end{table*}

\subsection{Human-Centered Evaluation of Generative AI}

Evaluating generative AI has traditionally relied on automated metrics to assess the quality of generated outputs. In image generation, commonly used measures include visual fidelity, text-image alignment, aesthetic quality, and learned preference scores. These metrics provide scalable and efficient evaluation mechanisms, but their improvements do not necessarily correspond to human perceptions or preferences. Prior research has identified discrepancies between automated metrics and human judgments, highlighting the importance of incorporating human evaluation when assessing text-to-image generation~\cite{otani2023toward,wiles2025revisiting}. From a human-centered perspective, evaluation should therefore consider not only measurable output quality but also whether generated results satisfy users' actual expectations.

HCI research has  demonstrated that evaluating generative systems requires attention to preferences and interaction contexts. For example, PromptCharm investigates how multimodal prompting and iterative refinement help users generate images that better match their intentions~\cite{wang2024promptcharm}. Similarly, RoomDreaming examines personalized design exploration by considering users' preferences, interaction efficiency, and satisfaction throughout the creative process~\cite{wang2024roomdreaming}. Collaborative fashion-design systems provide  HCI examples of generative AI embedded in multi-stage creative workflows. CrossGAI supports cross-device designer collaboration through AI-assisted sketch retrieval, sketch generation, and sketch-to-image synthesis, whereas HAIGEN integrates text-guided inspiration, personalized sketch-material retrieval, sketch refinement, and coloring \cite{deng2024crossgai,jiang2024haigen}. Other studies have shown that AI-generated content can influence users' creative exploration and decision-making, suggesting that improvements in generation capabilities do not necessarily lead to better interaction experiences~\cite{wadinambiarachchi2024effects}. These findings indicate that human-centered evaluation extends beyond assessing  generated outputs to understanding how users interact with, interpret, and refine generative results.

Meanwhile, the reliability of human evaluation remains a consideration. Human judgments may vary with evaluation criteria, participant backgrounds, and experimental protocols, making evaluation design and reproducibility essential~\cite{otani2023toward}. Accordingly, human-centered evaluation requires combining automated measurements with human judgments and contextual assessments to determine whether generative systems support user needs. This perspective provides an important foundation for examining how human-intent alignment is evaluated in diffusion model optimization.

% Required packages:
% \usepackage{booktabs}
% \usepackage{tabularx}
% \usepackage{array}
% \usepackage[table]{xcolor}

\begin{table*}[t]
\centering
\caption{Keyword search terms for identifying studies on human-centered diffusion model optimization.}
\label{tab:search_keywords}

\footnotesize
\renewcommand{\arraystretch}{1.15}
\setlength{\tabcolsep}{7pt}

\begin{tabularx}{\textwidth}{
    >{\raggedright\arraybackslash}p{0.095\textwidth}
    >{\raggedright\arraybackslash}X
}

\toprule
\rowcolor{gray!15}
\multicolumn{2}{l}{
\textbf{ACM Digital Library \& IEEE Xplore}
}\\
\midrule

\textbf{Title} &
\texttt{("diffusion model" OR "text-to-image diffusion" OR
"diffusion-based image generation")} \\

\addlinespace[3pt]

\textbf{Text} &
\texttt{("reinforcement learning" OR "RLHF" OR "DPO" OR
"preference optimization" OR "reward optimization" OR
"training-free guidance" OR "inference-time optimization" OR
"test-time optimization" OR "latent optimization" OR
"trajectory optimization")}
\par\medskip
\textbf{AND}
\par\medskip
\texttt{("human preference" OR "human feedback" OR
"human intent" OR "user preference" OR "alignment" OR
"personalization" OR "human-centered" OR "user interaction" OR
"interactive generation" OR "iterative refinement" OR
"human-AI co-creation")} \\

\midrule

\rowcolor{gray!15}
\multicolumn{2}{l}{
\textbf{Web of Science \& Scopus}
}\\
\midrule

\textbf{Title} &
\texttt{("diffusion model" OR "text-to-image diffusion" OR
"diffusion-based image generation")} \\

\addlinespace[3pt]

\textbf{Topic / Keywords} &
\texttt{("reinforcement learning" OR "RLHF" OR "DPO" OR
"preference optimization" OR "reward optimization" OR
"training-free guidance" OR "inference-time optimization" OR
"latent optimization" OR "trajectory optimization")}
\par\medskip
\textbf{AND}
\par\medskip
\texttt{("human preference" OR "human feedback" OR
"human intent" OR "user preference" OR "alignment" OR
"personalization" OR "human-centered" OR
"interactive generation" OR "user interaction")} \\

\midrule

\rowcolor{gray!15}
\multicolumn{2}{l}{
\textbf{arXiv}
}\\
\midrule

\textbf{Title / Abstract} &
\texttt{("diffusion model" OR "diffusion-based generation" OR
"text-to-image")}
\par\medskip
\textbf{AND}
\par\medskip
\texttt{("preference" OR "feedback" OR "alignment" OR
"reinforcement learning" OR "reward" OR "guidance" OR
"optimization" OR "personalization")} \\

\bottomrule
\end{tabularx}
\end{table*}

\begin{figure}
    \centering
    \includegraphics[width=1.0\linewidth]{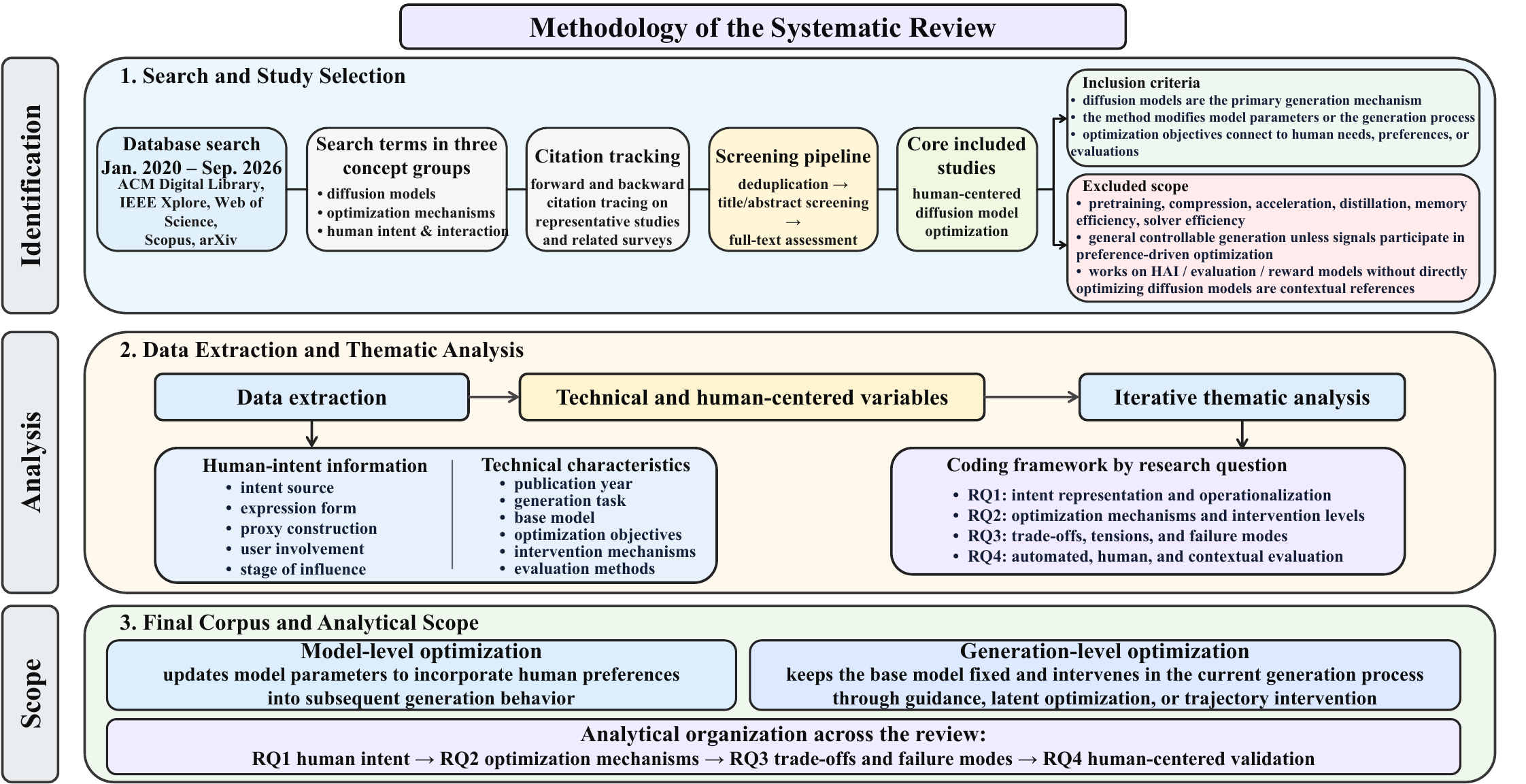}
    \caption{The review consists of three stages: \textbf{(1). }Identification and study selection, including database searches, keyword-based retrieval, citation tracking, screening, and the application of inclusion and exclusion criteria; \textbf{(2). }Data extraction and thematic analysis, which capture technical and human-centered variables and organize the extracted evidence according to four research questions; and \textbf{(3).} Final corpus and analytical scope, which distinguishes model-level from generation-level optimization and structures the subsequent synthesis around four interconnected dimensions: human intent representation and operationalization (RQ1), optimization mechanisms (RQ2), trade-offs and failure modes (RQ3), and human-centered validation (RQ4).}
    \label{Review_Methodology_Framework}
\end{figure}

\subsection{Existing Surveys and Our Perspective}

Existing surveys have examined diffusion models from several complementary perspectives. Early reviews primarily focused on the foundations, architectures, and generation capabilities of text-to-image diffusion models~\cite{zhang2023text}. Subsequent surveys extended this scope to controllable generation, systematically organizing existing methods according to control conditions, implementation mechanisms, and evaluation strategies~\cite{jiang2024survey,cao2025controllable}. More recently, research on diffusion model alignment has received increasing attention. Liu et al.~\cite{liu2026alignment} reviewed alignment objectives, preference modeling, optimization techniques, and evaluation benchmarks, while Wu et al.~\cite{wu2026preference} provided a comprehensive analysis of preference alignment in image generation and editing, covering reinforcement learning from human feedback, direct preference optimization, and reward modeling. Together, these surveys establish an important foundation for understanding the technical development of diffusion models and their alignment with human preferences.

Despite their complementary contributions, these reviews differ from our work in their primary analytical focus. \textbf{\textit{Existing surveys generally emphasize model capabilities, control mechanisms, optimization paradigms, or alignment techniques, whereas the transformation of human intent across different stages of optimization receives less integrated treatment. In particular, how user needs are expressed through interaction, translated into computational proxies, coordinated as optimization objectives, and subsequently validated against actual human preferences is not typically examined as a unified process.}} This distinction is important because successful optimization of a predefined objective does not necessarily imply that the resulting generation faithfully reflects the user's underlying intent, especially when preferences are subjective, multidimensional, and context-dependent.

\textbf{\textit{To complement existing surveys, our work adopts human intent transformation as its central analytical perspective.}} Rather than introducing another taxonomy based primarily on model architectures or optimization algorithms, we connect four interconnected dimensions: intent representation, optimization mechanisms, failure modes and trade-offs, and human-centered validation. This perspective enables us to examine how human needs enter diffusion model optimization, how computational objectives influence model behavior or generation trajectories, where deviations from human intent may emerge, and whether optimization outcomes are genuinely supported by human judgments. By integrating these dimensions into a unified analytical framework, our review provides a human-centered understanding of diffusion model optimization that complements existing technical and preference-alignment surveys.
\textbf{Tab}.~\ref{tab:survey_comparison} compares representative surveys on diffusion models, controllable generation, and preference alignment with our human-centered perspective.
Unlike prior surveys primarily organized around generation capabilities, control mechanisms, or alignment techniques, our review examines how human intent is represented, transformed into computational objectives, executed through optimization, potentially distorted, and ultimately validated against actual human needs.

\begin{figure}
    \centering
    \includegraphics[width=1.0\linewidth]{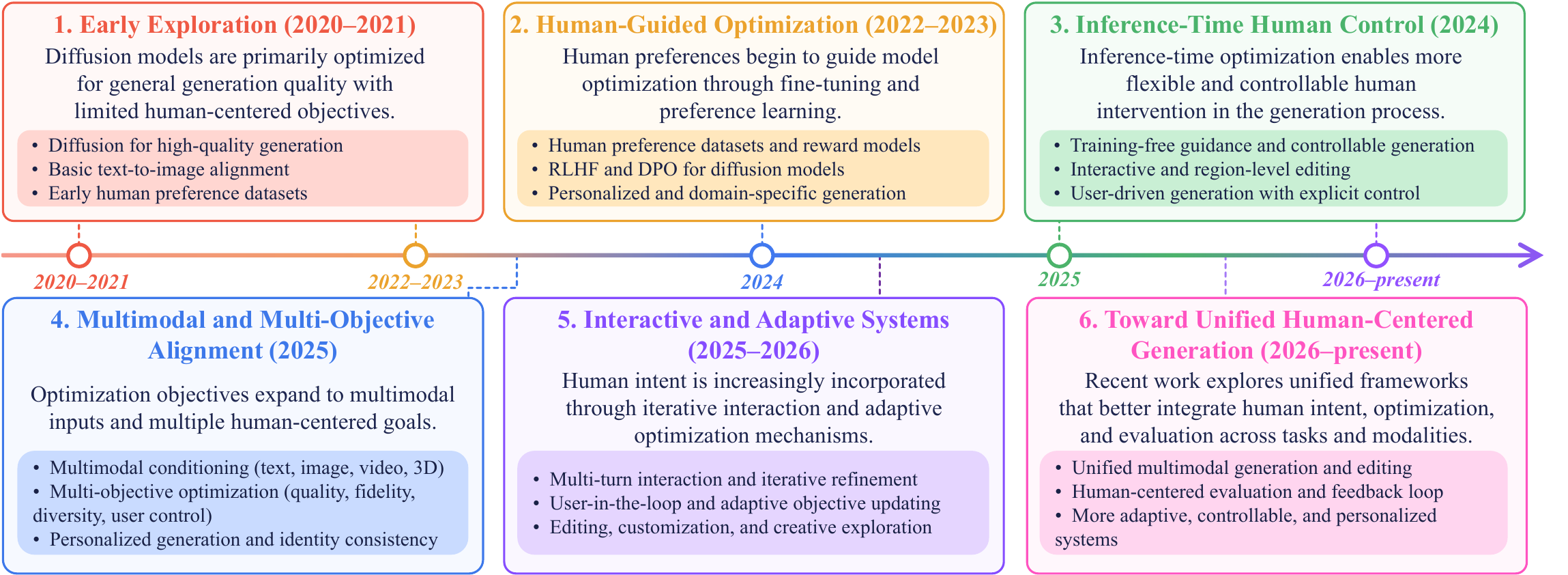}
    \caption{Evolution of human-centered diffusion model optimization. The timeline illustrates six successive research stages: \textbf{(1).} early exploration of diffusion models with limited human-centered objectives;\textbf{ (2).} human-guided optimization through fine-tuning and preference learning; \textbf{(3).} inference-time human control and controllable generation; \textbf{(4).} multimodal and multi-objective alignment; \textbf{(5).} interactive and adaptive optimization systems; and \textbf{(6).} emerging unified human-centered generation frameworks. Representative models and methods are included to illustrate the major developments and shifting research priorities across these stages.}
    \label{fig:timeline}
\end{figure}

\section{Review Methodology}
\label{sec:Review Methodology}

To systematically examine how human intent enters and shapes diffusion model optimization, we conduct a systematic literature review guided by the four research questions. Our methodology consists of three stages: \textbf{\textit{First}}, we identify and screen relevant studies through cross-database searches, citation tracking, and predefined selection criteria; \textbf{\textit{Second}}, we extract technical and human-centered information from the selected studies and conduct iterative thematic analysis; \textbf{\textit{Finally}}, we organize the resulting evidence around four interconnected dimensions: intent representation, optimization mechanisms, failure modes and trade-offs, and human-centered validation. An overview of the systematic review workflow, including study identification, thematic analysis, and the final analytical scope, is presented in \textbf{Fig}.~\ref{Review_Methodology_Framework}.

\subsection{Search and Study Selection}

We searched studies published between January 2020 and September 2026 on diffusion model optimization, human preferences, and human--AI interaction. The databases included ACM Digital Library, IEEE Xplore, Web of Science, Scopus, and arXiv, covering HCI, computer vision, machine learning, and generative AI.

The search strategy was organized around three concept groups, as summarized in \textbf{Tab.~\ref{tab:search_keywords}}.
The search terms were organized into three concept groups. The first focused on diffusion models, including \textit{diffusion model}, \textit{text-to-image diffusion}, and \textit{diffusion-based image generation}. The second identified optimization mechanisms, including \textit{reinforcement learning}, \textit{RLHF}, \textit{preference optimization}, \textit{DPO}, \textit{reward optimization}, \textit{training-free guidance}, \textit{inference/test-time optimization}, \textit{latent optimization}, and \textit{trajectory optimization}. The third focused on human intent and interaction, including \textit{human preference}, \textit{human feedback}, \textit{human intent}, \textit{user preference}, \textit{alignment}, \textit{personalization}, \textit{human-centered}, \textit{user interaction}, \textit{interactive generation}, \textit{iterative refinement}, and \textit{human-AI co-creation}. Since different research communities describe human-centered optimization using different terminology, we did not require studies to explicitly mention \textit{human intent} or \textit{human-centered}. Instead, we examined whether their optimization objectives could be meaningfully connected to human needs, preferences, or interaction signals during full-text screening.

To reduce omissions caused by keyword variation, we further conducted forward and backward citation tracking on representative studies of reinforcement-learning fine-tuning, preference optimization, training-free guidance, and diffusion alignment, and examined references in related surveys. All studies identified through citation tracking were screened using the same criteria as database results.

The selection process included deduplication, title--abstract screening, and full-text assessment. We excluded surveys, opinion articles, tutorials, posters, theses, and short papers without complete method descriptions from the core research corpus; relevant surveys were retained for background analysis and citation tracking. During full-text screening, studies were assessed according to our definition of \textit{human-centered diffusion model optimization}. Included studies had to satisfy two criteria. First, diffusion models had to serve as the primary generation mechanism, with the proposed method modifying model parameters or the generation process. Second, the optimization objectives had to be directly or indirectly connected to human needs, preferences, or evaluations. Such connections could be established through explicit user feedback, ratings, rankings, pairwise preferences, or computational proxies, including reward and preference models trained on human evaluation data~\cite{zhong2025comprehensive}. Importantly, these different forms of human involvement were not treated as equivalent, since computational proxies may capture only selected aspects of the original human intent.

Accordingly, we excluded studies focused solely on pretraining, model compression, sampling acceleration, distillation, memory efficiency, or numerical solver efficiency. General conditional or controllable generation was not included simply because it employed textual, positional, or structural conditions, unless these signals further participated in preference-driven optimization. Studies focusing on human--AI interaction, user experience, intent expression, or evaluation without directly optimizing diffusion models were retained as contextual references rather than included in the core corpus. Similarly, works proposing reward models, evaluation metrics, or benchmarks without optimizing diffusion models were used as supplementary evidence for analyzing human-centered evaluation practices.

\subsection{Data Extraction and Thematic Analysis}

For each included study, we extracted information related to human intent, including intent source, expression form, proxy construction, user involvement, and the stage at which human preferences influenced optimization. We also recorded technical characteristics, including publication year, generation task, base model, optimization objectives, intervention mechanisms, and evaluation methods. These variables allowed us to examine both how human needs were operationalized and how the resulting objectives shaped diffusion model optimization.

The coding framework was organized around the four research questions. For RQ1, we examined how human intent was expressed, represented, and translated into rewards, preference signals, or other computational proxies. For RQ2, we analyzed how these objectives influenced model behavior or generation trajectories, distinguishing \textit{model-level} and \textit{generation-level optimization}. For RQ3, we recorded proxy-objective mismatches, competing objectives, diversity-related trade-offs, optimization costs, and reported failure modes. For RQ4, we extracted automated metrics, human judgments, expert evaluations, and contextual validation methods, with particular attention to whether evaluation practices reflected individual preferences and actual user needs.

We then conducted iterative thematic analysis to identify recurring patterns and differences across studies. Rather than organizing the literature solely by algorithm type, we compared studies according to how human intent was represented, how optimization objectives were applied, what tensions emerged, and how alignment outcomes were evaluated. Related codes were grouped into broader themes, while distinctions in intent sources, proxy construction, and optimization intervention points were preserved. Ambiguous cases were revisited to refine the coding categories and maintain consistent interpretation across the research corpus.

\subsection{Final Corpus and Analytical Scope}

The final research corpus covered two main pathways through which human-centered objectives influence diffusion model optimization. \textit{Model-level optimization} updates model parameters to incorporate human preferences into generation behavior, enabling relatively persistent adaptation across subsequent generations. In contrast, \textit{generation-level optimization} keeps the base model fixed and adjusts the current generation process through guidance, latent optimization, or trajectory intervention. These pathways differ not only in their optimization mechanisms but also in how human intent influences generation behavior and when users can intervene. Since some studies involved multiple mechanisms, we categorized them according to their primary intervention points and the ways in which optimization objectives were applied, without treating the two pathways as mutually exclusive.

Rather than organizing the corpus solely by algorithm type, we structured the analysis around the four research questions. Specifically, we first examine how human intent is expressed, represented, and operationalized into computational objectives (RQ1). We then analyze how these objectives influence model behavior or generation trajectories through different optimization mechanisms (RQ2). Next, we investigate the trade-offs, tensions, and failure modes that may emerge when optimizing toward human-defined or human-proxied objectives (RQ3). Finally, we examine how automated and human-centered evaluation practices validate whether optimization outcomes genuinely align with human preferences and user needs (RQ4). Together, these dimensions provide a unified analytical scope for understanding how human intent is represented, executed, potentially distorted, and evaluated throughout diffusion model optimization. To provide an overview of the analytical structure developed in the following sections, \textbf{Tab}.~\ref{tab:proposed_taxonomy_tables} summarizes the proposed taxonomy tables across Sections~4--9. It outlines the main classification dimensions covered in each section, ranging from the stakeholders shaping optimization goals and the operationalization of human intent to optimization methods, associated trade-offs, evaluation practices, and future research directions. This overview provides a roadmap for the subsequent systematic analysis.

% \textcolor{red}{To characterize the literature coverage of the analytical sections, we counted distinct bibliography entries within each section, counting repeated citations only once per section. Section.~\ref{Landscape} contains 11 distinct references related to the research landscape; Section.~\ref{Human_Intent_to_Optimization} contains 22 related to intent representation and objective operationalization; Section.~\ref{Optimizing_Diffusion_Models}  contains 36 related to optimization mechanisms; Section.~\ref{section7}  contains 5 related to tensions and trade-offs; and Section.~\ref{section8} contains 28 related to evaluation practices. These counts sum to 102 section-level reference occurrences, with studies cited across multiple sections counted separately. They describe the distribution of references across the analytical sections rather than the number of unique studies included in the core research corpus.}

\section{The Landscape of Human-Centered Diffusion Model Optimization}
\label{Landscape}

\begin{figure}
    \centering
    \includegraphics[width=1.0\linewidth]{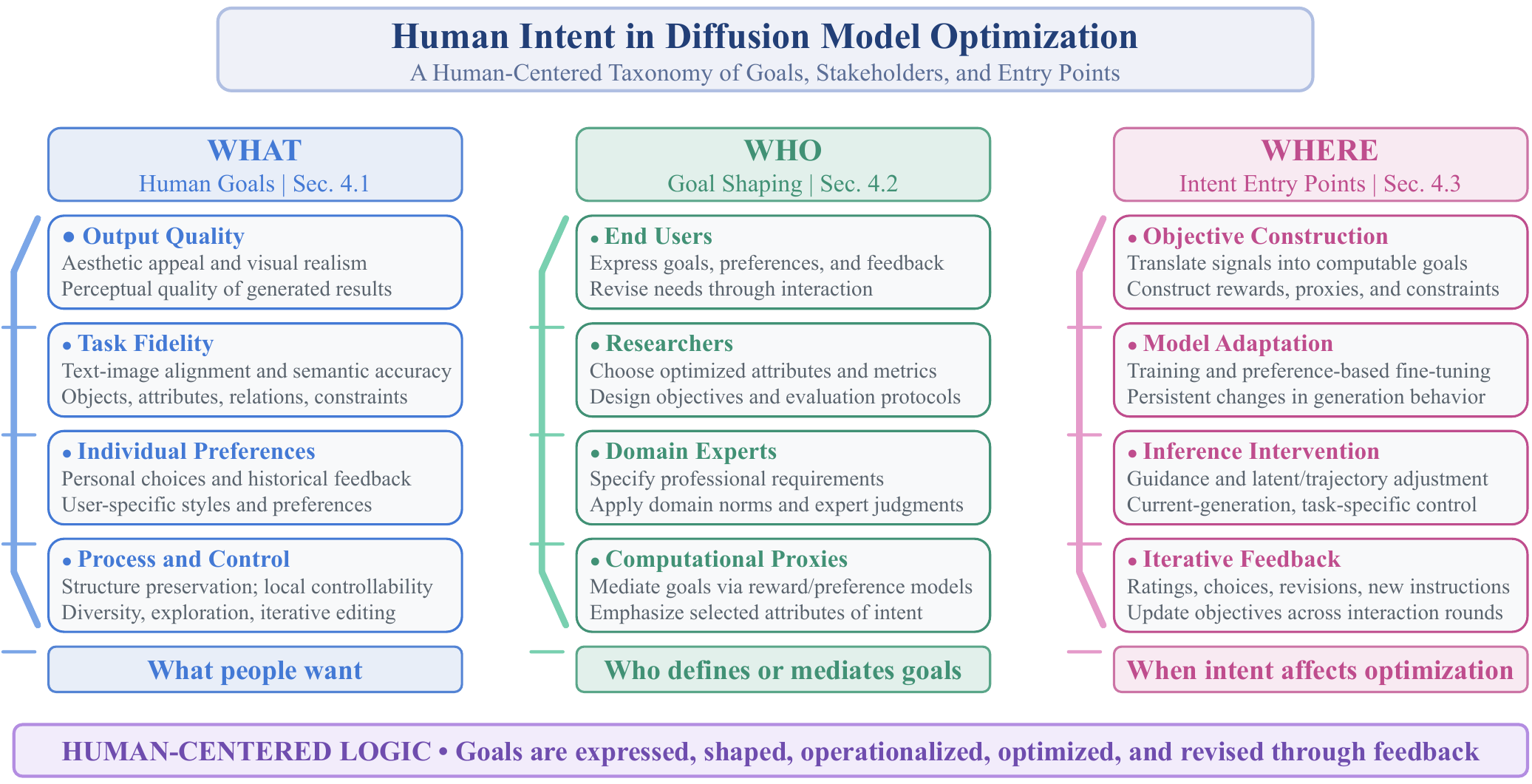}
    \caption{A human-centered taxonomy of human intent in diffusion model optimization, organized along three dimensions: \textbf{(1)} What: the goals that human-centered optimization seeks to satisfy, including output quality, task fidelity, individual preferences, and process control; \textbf{(2)} Who: the stakeholders who define or mediate optimization goals, including end users, researchers, domain experts, and computational proxies; and \textbf{(3)} Where: the entry points through which human intent influences optimization, including objective construction, model adaptation, inference-time intervention, and iterative feedback. Together, these dimensions characterize how human intent is expressed, shaped, operationalized, and continuously refined throughout diffusion model optimization.}
    \label{fig:section4what_who}
\end{figure}

% \begin{figure}
%     \centering
%     \includegraphics[width=1\linewidth]{sec4.png}
%         \vspace{-0.8cm}
%     \caption{This paper characterizes the research space of human-centered diffusion model optimization along three dimensions: what human generation goals existing work seeks to satisfy, who actually shapes the optimization goals, and at which stages human intent enters the generation and optimization process.}

%     \label{fig:human-centered-landscape}
% \end{figure}

This section characterizes the research landscape of human-centered diffusion model optimization through three interconnected dimensions derived from our thematic analysis. First, we examine the diverse human goals that existing optimization methods seek to satisfy, including generation quality, task fidelity, and individual preferences. Second, we analyze who defines and shapes these optimization goals, considering the roles of users, researchers, domain experts, and computational proxies. Third, we investigate how and at which stages human intent enters the optimization process through different interaction signals. Together, these dimensions provide a human-centered understanding of the research landscape and establish the foundation for the next section, which further examines how human intent is represented and translated into machine-optimizable objectives. \textbf{Fig}.~\ref{fig:section4what_who} summarizes these three dimensions, providing an integrated taxonomy of human goals, the stakeholders shaping optimization objectives, and the entry points through which human intent influences the optimization process.

\subsection{What Human Goals Are Targeted in Diffusion Optimization?}

Existing studies often describe their goals as `better meeting human needs', but these needs encompass different dimensions of human intent. Some methods primarily focus on the quality of generated outputs, such as aesthetic appeal, visual realism, or perceptual quality. Others emphasize how faithfully the model responds to user requests, including text-image alignment and the accurate depiction of objects, attributes, and relations. Thus, even when described as alignment, different methods may prioritize aesthetic preferences or task fidelity~\cite{huang2025smartspatial,li2024stylegan}.

With the development of personalized generation, optimization goals have increasingly expanded beyond group-level preferences to account for individual differences. The central question is therefore not only `what kind of image is generally better', but also `what kind of result better satisfies the needs of a particular user'. Some studies incorporate personal choices, historical feedback, or repeated interactions to adapt generation behavior to individual preferences~\cite{li2025beyond,zhao2025catversion,xu2024sgdm,xiao2025fastcomposer,peng2024portraitbooth}.

Human needs extend beyond the quality of the final image. In interactive creation and editing, users may value structure preservation, local controllability, output diversity, and opportunities for creative exploration. These aspects influence not only what users expect from generated results but also how they wish to interact with and control the generation process. As summarized in \textbf{Tab}.~\ref{tab:human_intent_taxonomy}, human intent in diffusion model optimization encompasses four categories: output quality, task fidelity, individual preferences, and process-level control. These categories are not mutually exclusive; they may coexist within a task and evolve in relative importance as users interact with the system.

\subsection{Who Shapes the Optimization Goal?}

\begin{table*}[t]
\caption{Taxonomy of human goals targeted in diffusion model optimization. These categories are analytically distinguishable but often co-exist within a single generation or editing task.}
\label{tab:human_intent_taxonomy}
\centering
\small
\setlength{\tabcolsep}{4pt}
\renewcommand{\arraystretch}{1.12}
\begin{tabularx}{\textwidth}{Y{0.82} Y{0.92} Y{1.08} Y{1.18}}
\toprule
\textbf{Intent category} & \textbf{Core concern} & \textbf{Typical optimization goals} & \textbf{Representative user needs / signals} \\
\midrule

\textbf{Quality-Oriented}
& Improving the perceived quality of generated results
& Aesthetic appeal, visual realism, perceptual quality, image attractiveness
& ``Make it look better,'' ``more realistic,'' ``more beautiful,'' or preferences reflected in ratings or pairwise judgments \\
\addlinespace

\textbf{Fidelity-Oriented}
& Ensuring that outputs faithfully satisfy the user’s request or task specification
& Text--image alignment, accurate depiction of objects, attributes, relations, and structural constraints
& ``Follow my prompt more accurately,'' ``preserve the object layout,'' ``edit only this part,'' or instruction-following feedback \\
\addlinespace

\textbf{Preference-Oriented}
& Adapting generation to differences across users rather than optimizing only group-level averages
& Personalized style preference, user-specific aesthetics, historical preference adaptation, identity-consistent generation
& Personal choices, historical interactions, repeated feedback, user profiles, or multi-round preference refinement \\
\addlinespace

\textbf{Control-Oriented}
& Supporting how users interact with and steer the generation process, not only the final result
& Structure preservation, local controllability, output diversity, exploration support, editability, interactive refinement
& Region selection, reference images, iterative edits, requests for multiple alternatives, and needs for controllable exploration \\
\bottomrule
\end{tabularx}
\end{table*}

Optimization goals in diffusion models are not always directly defined by end users. In user-driven settings, individuals express their needs through text, ratings, rankings, or preference comparisons, allowing their judgments to influence the optimization direction. However, the extent of this influence depends on how these signals are interpreted and incorporated into the optimization process.

In many studies, researchers determine which attributes should be optimized, such as aesthetic quality, text-image alignment, or task-specific performance. Human feedback is then collected or interpreted within these predefined dimensions. Consequently, even when optimization relies on human evaluations, the underlying definition of what constitutes a better result may still be largely determined by the research design. Similarly, in professional domains such as medicine or design, domain experts and established professional standards may influence the definition of desirable outputs. In these settings, optimization goals can reflect both individual user needs and domain-specific requirements.

Computational proxies introduce a layer of mediation between human judgments and optimization objectives. Reward models, preference models, and VLM-based evaluators can transform human evaluations or predefined criteria into automatic optimization signals. However, these proxies may preserve certain aspects of human preferences while overlooking others, thereby influencing which objectives are reinforced during optimization~\cite{huang2025patchdpo}. Thus, optimization goals may be shaped through the combined influence of users, researchers, domain experts, and computational proxies, although their roles in defining and mediating these goals differ. Distinguishing these roles is essential for understanding whose preferences are represented and how faithfully the objectives reflect the intentions of specific users.

\subsection{Where Does Human Intent Enter the Optimization Process?}

Human intent can be expressed through various interaction signals, including text prompts, ratings, rankings, pairwise preferences, region selection, reference images, multi-turn edits, interaction history, and expert judgments in professional settings. These signals differ in their information granularity, user effort, and ability to communicate specific preferences or constraints. However, not every interaction signal directly participates in optimization. Within the scope of this review, such signals become relevant to diffusion model optimization when they are used to construct objectives, guide model adaptation, or actively intervene in the generation process.

\begin{figure}
    \centering
    \includegraphics[width=1.0\linewidth]{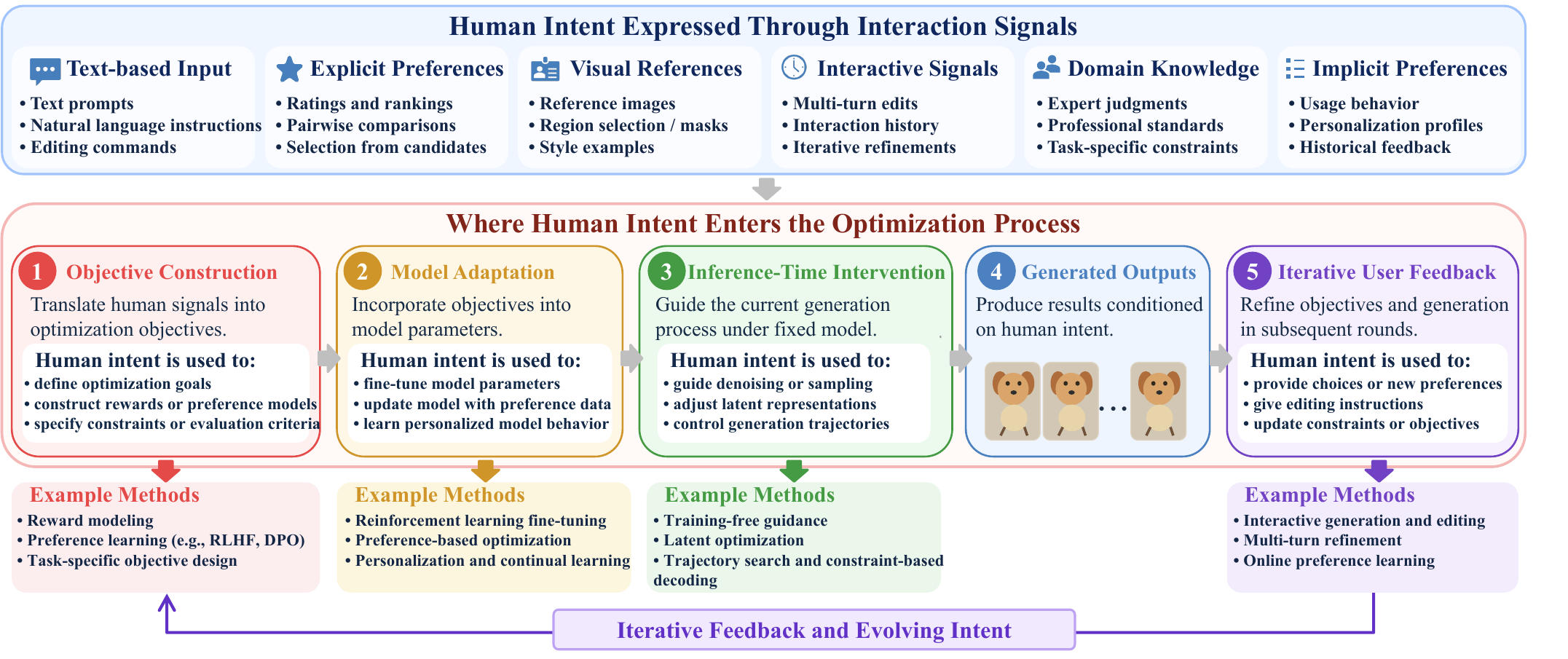}
    \caption{Human intent and its entry points in diffusion model optimization. Human needs are expressed through diverse interaction signals and can influence objective construction, model adaptation, inference-time intervention, and iterative feedback. These entry points are not mutually exclusive and may jointly support evolving user preferences across generation and editing tasks.}
    \label{CHI_workflow}
\end{figure}

Human intent can influence optimization at different stages rather than entering the system only through an initial request. Some studies incorporate human preferences during feedback collection and objective construction, translating user judgments into rewards, preference models, or other computational proxies. Others introduce these objectives during training or fine-tuning to influence subsequent generation behavior. Alternatively, human-defined or human-proxied objectives can guide the current generation process at inference time through latent adjustments, guidance mechanisms, or trajectory interventions~\cite{ding2025dis2booth,aiello2025dreamcache,huang2025resolving}. After generation, users may further provide choices, ratings, or editing instructions that inform subsequent rounds of interaction and optimization.  \textbf{Fig}.~\ref{CHI_workflow} provides an overview of the interaction signals through which human intent is expressed and illustrates how these signals influence different stages of the diffusion model optimization process, from objective construction and model adaptation to inference-time intervention and iterative user feedback.

Therefore, human intent should not be understood as a fixed condition supplied only at the system input. Instead, it may enter and influence diffusion model optimization through objective construction, model adaptation, inference-time intervention, and iterative user feedback. Identifying these entry points helps clarify when users can influence generation behavior and provides a basis for examining how their expressed intentions are subsequently transformed into machine-optimizable objectives.

\section{From Human Intent to Optimization Objectives}
\label{Human_Intent_to_Optimization}

% \begin{figure*}
%     \centering
%     \includegraphics[width=1\linewidth]{sec5.png}
%     \vspace{-0.6cm}
%     \caption{This figure summarizes how human intent is progressively transformed into optimizable objectives. User needs are first represented as machine-processable attributes or preference signals and are then further constructed as proxy objectives. However, as user preferences change across tasks and generated results, fixed objectives may no longer continuously reflect current needs, ultimately leading to a Human–Proxy Gap.}
%     \label{fig:intent-to-objectives}
% \end{figure*}

Sec.~\ref{Landscape} has characterized what human goals existing studies seek to satisfy, who shapes these goals, and where human intent enters diffusion model optimization. Building on this analysis, this section examines how expressed human needs are represented and operationalized into computational objectives, thereby addressing RQ1. Specifically, we trace the transformation from intent representation and proxy construction to objective coordination and potential proxy mismatches, examining how human intent becomes machine-optimizable and where information may be lost or distorted. \textbf{Tab}.~\ref{tab:intent_operationalization_taxonomy} summarizes the main categories of human intent representation and operationalization identified in our review. Specifically, we distinguish intent representation, proxy objective construction, objective coordination, and proxy mismatch, highlighting how human needs are progressively transformed into machine-optimizable objectives and where discrepancies may arise. This classification provides an analytical overview of RQ1 and establishes the foundation for the following discussion.

\subsection{From Expressed Intent to Machine-Readable Representations}

Human needs, particularly those that are subjective or context-dependent, cannot always be directly used as computational optimization objectives. Once expressed through interaction signals, human intent must be formalized into attributes, constraints, or preferences that can be represented, compared, and computed. For example, `better matching my taste' may be represented through aesthetic quality, visual realism, or style preferences, while `better following my request' may be expressed through semantic consistency, object attributes, or structural constraints~\cite{xing2024svgdreamer}.

This formalization determines what information subsequent optimization mechanisms can access. Explicit requirements can often be preserved as conditions or target attributes, whereas subjective preferences may require ratings, rankings, or pairwise comparisons to become computationally interpretable. More complex needs may also be represented through multimodal inputs, interaction history, or personalized user models~\cite{lu2025inpo,dang2025personalized,li2026instant}.

Therefore, intent representation is not simply the preservation of user input, but a selective transformation of human needs into machine-readable information. While this process enables computational optimization, it also determines which aspects of the original intent are retained and which may be omitted before subsequent optimization begins.

\begin{table*}[t]
\centering
\caption{Taxonomy of Human Intent Representation and Operationalization. Existing studies and analytical concepts are organized according to how human needs are represented, transformed into computational objectives, coordinated, and potentially misaligned during optimization.}
\label{tab:intent_operationalization_taxonomy}

\small
\renewcommand{\arraystretch}{1.25}
\setlength{\tabcolsep}{5pt}

\begin{tabularx}{\textwidth}{
    >{\raggedright\arraybackslash}p{0.16\textwidth}
    >{\raggedright\arraybackslash}p{0.21\textwidth}
    >{\raggedright\arraybackslash}X
    >{\raggedright\arraybackslash}p{0.18\textwidth}
}
\toprule
\textbf{Category} &
\textbf{Sub-topic} &
\textbf{Description} &
\textbf{References} \\
\midrule

\multirow{2}{*}{\parbox{0.16\textwidth}{
\raggedright Intent Representation
}}
&
Explicit Requirements and Constraints &
Formalizes expressed needs as semantic attributes,
structural constraints, or task conditions. &
\cite{xing2024svgdreamer} \\

\addlinespace[5pt]

&
Multimodal and Personalized Representations &
Uses multimodal information, interaction histories,
or personalized representations to encode user needs. &
\cite{lu2025inpo,dang2025personalized,li2026instant} \\

\midrule

\multirow{2}{*}{\parbox{0.16\textwidth}{
\raggedright Proxy Objective Construction
}}
&
Reward and Preference Modeling &
Converts human judgments into reusable reward functions
or preference-based objectives. &
\cite{lee2025calibrated,yuan2024instructvideo,ye2025schedule,wang2025align,huang2025dreamfuse,luo2025adding,liang2024rich,xing2025svgdreamer,croitoru2025curriculum,li2025reneg,wang2025mvreward} \\

\addlinespace[5pt]

&
Predefined Computational Evaluators &
Employs predefined evaluation criteria or learned
computational representations as optimization proxies. &
\cite{yang2026optimizing} \\

\midrule

Objective Coordination &
Multi-Objective Aggregation &
Coordinates potentially competing objectives involving
visual quality, fidelity, diversity, and user preferences. &
\cite{xie2025dymo2,yan2026explore,yanmultitune} \\

\midrule

Proxy Mismatch &
Human--Proxy Gap &
Characterizes potential information loss or misalignment
between human needs and their computational representations. &
\cite{karthik2025scalable,liu2025dreamreward,yang2024human} \\

\bottomrule
\end{tabularx}
\end{table*}

\subsection{From Human Signals to Proxy Objectives}

Machine-readable representations of human intent do not necessarily constitute objectives that optimization algorithms can directly optimize. Whether these representations originate from textual instructions, preference judgments, or other interaction signals, they often need to be translated into computational objectives that can evaluate generated results and provide optimization signals. This translation forms a critical connection between human intent expression and diffusion model optimization.

One common approach constructs reward or preference models from human evaluations, allowing discrete judgments to be approximated by reusable  functions~\cite{lee2025calibrated,yuan2024instructvideo,ye2025schedule,wang2025align,huang2025dreamfuse,luo2025adding,liang2024rich,xing2025svgdreamer,croitoru2025curriculum,li2025reneg,wang2025mvreward}. Approaches employ predefined aesthetic metrics, vision-language models, or task-specific evaluators to approximate selected aspects of human needs as scores. Although these objectives differ in their sources and construction methods, they serve a similar function in optimization: providing measurable signals that guide model updates or generation-time interventions.

This operationalization improves the computability and scalability of human-centered objectives, but it also introduces an additional layer between human intent and algorithmic optimization. Rather than directly optimizing the original human need, algorithms operate on proxy objectives derived from human judgments, learned representations, or predefined evaluation criteria~\cite{yang2026optimizing}. Consequently, proxy construction determines not only which objectives can be optimized, but also which aspects of human intent are emphasized or overlooked during subsequent generation.

\subsection{Multi-Objective Trade-offs and Evolving Intent}

After human needs are translated into computational objectives, a further challenge arises from the relationships among these objectives. Personalized generation often involves multiple dimensions, including aesthetic quality, semantic consistency, realism, diversity, and structure preservation. However, these objectives cannot always be improved simultaneously. For example, stronger semantic constraints may restrict the available exploration space, while emphasizing aesthetic rewards may favor a narrower range of high-scoring outputs~\cite{xie2025dymo2,yan2026explore,yanmultitune}. Optimization systems often employ weighting, ranking, or other aggregation strategies to coordinate competing objectives.

A central difficulty is that objective relationships are frequently established through relatively static weights or predefined priorities, whereas human preferences may evolve across tasks and interactions. For instance, users may initially prioritize semantic fidelity but later place greater importance on diversity after observing several generated candidates. Similarly, during image editing, structure preservation may become increasingly important as an output approaches the desired result. Such changes mean that a fixed combination of objectives may no longer accurately reflect the user's current priorities.

Therefore, multi-objective coordination is not merely a technical problem of optimizing several metrics simultaneously. It also concerns whether the relationships among computational objectives remain consistent with the user's evolving intent. The way objectives are combined effectively determines what the optimization system considers a better result. When these relationships fail to adapt to changing human preferences, the constructed optimization objective may gradually diverge from what the user actually seeks to achieve.

\subsection{The Human--Proxy Gap}

Based on the preceding analysis, we characterize the potential structural mismatch between human intent and the computational objectives used by optimization algorithms as the \textit{Human--Proxy Gap}. Before human needs can guide optimization, they undergo several transformations, including intent expression, machine-readable representation, proxy construction, and objective coordination. Each transformation may preserve certain aspects of human intent while omitting contextual information, individual differences, or relationships among objectives~\cite{liu2025dreamreward}. Consequently, the resulting proxy objectives may represent only part of what users actually want.

Importantly, the Human--Proxy Gap is not equivalent to reward prediction error~\cite{karthik2025scalable}. Even when a reward model accurately predicts its predefined evaluation target, the optimization system may still produce unsatisfactory results if that target does not fully capture the user's needs, if competing objectives are combined inappropriately, or if preferences change over time. In such cases, the proxy objective may be accurately estimated and successfully optimized while remaining misaligned with the user's current intent~\cite{yang2024human}.

From this perspective, addressing RQ1 requires more than identifying which rewards or preference signals are used in existing methods. It requires understanding how human needs are progressively transformed into computational objectives and where discrepancies may emerge during this process. The Human--Proxy Gap therefore provides a foundation for the analysis of optimization mechanisms. The next section examines how model-level and generation-level optimization execute these constructed objectives to influence model behavior or generation trajectories.

\section{Optimizing Diffusion Models Toward Human Intent}
\label{Optimizing_Diffusion_Models}

% \begin{figure}
%     \centering
%     \includegraphics[width=1\linewidth]{sec6.png}
%     \caption{This figure compares how human intent is executed through two diffusion model optimization pathways. Model-level adaptation updates model parameters to persistently incorporate user preferences into subsequent generation behavior, whereas generation-level steering keeps the base model fixed and dynamically adjusts the generation trajectory according to the objective during inference. The two mechanisms also provide different spaces for user intervention: the former emphasizes persistent adaptation, while the latter preserves more direct and immediate control.}
%     \label{fig:intent-optimization}
% \end{figure}

\begin{figure}
    \centering
    \includegraphics[width=1.0\linewidth]{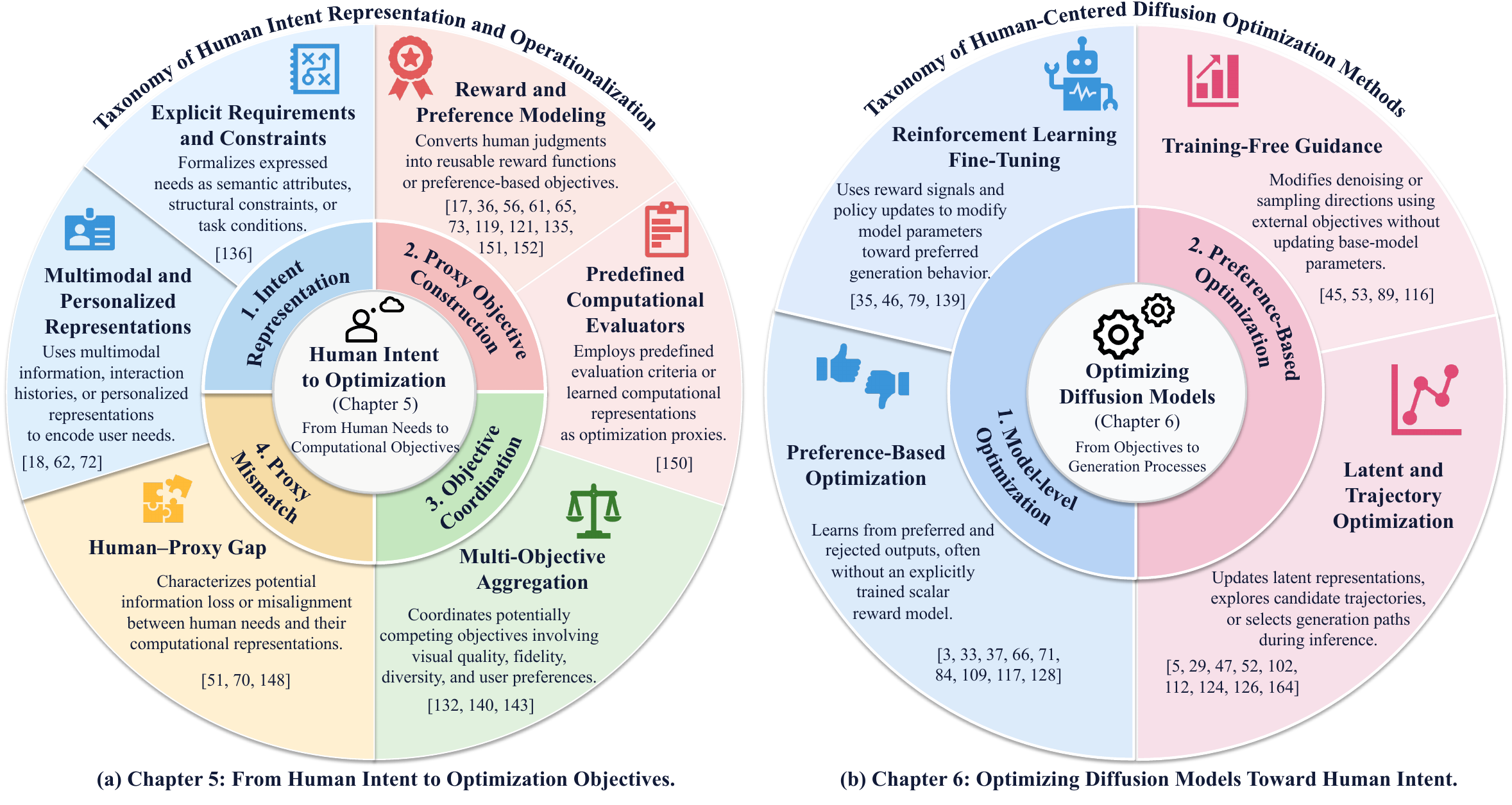}
    \caption{Taxonomies of human-intent operationalization and human-centered diffusion model optimization. \textbf{(a) }From Human Intent to Optimization Objectives : A taxonomy of how human needs are represented and translated into computational objectives, covering intent representation, proxy objective construction, objective coordination, and human–proxy mismatch. \textbf{(b)} From Objectives to Generation Processes : A taxonomy of human-centered diffusion optimization methods, distinguishing model-level optimization, including reinforcement learning fine-tuning and preference-based optimization, from generation-level optimization, including training-free guidance and latent/trajectory optimization. Together, the two taxonomies illustrate how human needs are transformed into computational objectives and subsequently operationalized through different diffusion optimization mechanisms.}
    \label{fig:dual_taxonomy}
\end{figure}

This section addresses RQ2 introduced in Sec.~\ref{sec:intro}. Sec.~\ref{Human_Intent_to_Optimization} has examined how human needs are represented and transformed into machine-optimizable proxy objectives. Building on this analysis, we investigate how these objectives are applied to diffusion models and translated into changes in generation behavior. Existing methods can be broadly organized into two pathways: \textit{model-level optimization}, which updates model parameters to influence subsequent generations, and \textit{generation-level optimization}, which keeps the base model fixed while intervening in the current generation process. Beyond comparing their technical mechanisms, we examine how these pathways differ in the persistence of preference adaptation and the opportunities they provide for user intervention, thereby connecting diffusion model optimization with human agency.
\textbf{Tab}.~\ref{tab:optimization_taxonomy} summarizes the main categories of human-centered diffusion model optimization methods, distinguishing model-level and generation-level approaches according to their optimization mechanisms and intervention points.To provide a consolidated view of the two analytical dimensions, \textbf{Fig}.~\ref{fig:dual_taxonomy} presents complementary taxonomies developed in Sections 5 and 6. The left panel organizes human intent operationalization into four dimensions: intent representation, proxy objective construction, objective coordination, and proxy mismatch. It further identifies representative mechanisms within each dimension, including explicit requirements, personalized representations, reward modeling, computational evaluators, and multi-objective aggregation. The right panel categorizes human-centered diffusion optimization methods into two primary paradigms: model-level optimization, which includes reinforcement learning fine-tuning and preference-based optimization, and generation-level optimization, which encompasses training-free guidance and latent or trajectory optimization. Together, these taxonomies connect the construction of machine-optimizable objectives with their subsequent execution in diffusion models, providing a unified view of how human intent is translated into and acted upon by optimization mechanisms.

\subsection{Model-Level: Persistent Intent Adaptation}

Model-level optimization updates model parameters according to constructed proxy objectives, allowing human preferences to influence subsequent generation behavior beyond a single inference process. Representative approaches include reinforcement learning fine-tuning and preference-based optimization. Although these approaches differ in how preference information is represented and incorporated into training, both aim to adjust the model's generation distribution toward outputs favored by their respective objectives.

\begin{table*}[t]
\centering
\caption{Taxonomy of Human-Centered Diffusion Model Optimization Methods. Existing approaches are categorized according to their primary intervention level and optimization mechanism.}
\label{tab:optimization_taxonomy}

\small
\renewcommand{\arraystretch}{1.25}
\setlength{\tabcolsep}{5pt}

\begin{tabularx}{\textwidth}{
    >{\raggedright\arraybackslash}p{0.17\textwidth}
    >{\raggedright\arraybackslash}p{0.19\textwidth}
    >{\raggedright\arraybackslash}X
    >{\raggedright\arraybackslash}p{0.20\textwidth}
}
\toprule
\textbf{Category} &
\textbf{Sub-topic} &
\textbf{Description} &
\textbf{Examples} \\
\midrule

\multirow{2}{*}{\parbox{0.17\textwidth}{
\raggedright Model-Level Optimization
}}
&
Reinforcement Learning Fine-Tuning &
Uses reward signals and policy updates to modify model parameters toward preferred generation behavior. &
\cite{yan2025entropy2,hu2025better,jia2025reward,miao2024training} \\

\addlinespace[5pt]

&
Preference-Based Optimization &
Learns from preferred and rejected outputs, often without an explicitly trained scalar reward model. &
\cite{liang2025aesthetic,wu2025rethinking,na2025boost,liu2025videodpo,huang2026dreamfuse,sun2025generalizing,hong2026margin,an2025agfsync,wang2026self} \\

\midrule

\multirow{2}{*}{\parbox{0.17\textwidth}{
\raggedright Generation-Level Optimization
}}
&
Training-Free Guidance &
Modifies denoising or sampling directions using external objectives without updating base-model parameters. &
\cite{wallace2023end,park2025steering,jeong2025stylekeeper,kim2025free2} \\

\addlinespace[5pt]

&
Latent and Trajectory Optimization &
Updates latent representations, explores candidate trajectories, or selects generation paths during inference. &
\cite{wei2026enhancing,wu2026noisepo,kim2025text,um2025minority,bahng2025cycle,zhou2025scale,simon2025titan,jiang2025mc2,guo2024focus} \\

\bottomrule
\end{tabularx}
\end{table*}

Reinforcement learning fine-tuning typically evaluates generated outputs using reward signals and updates model parameters to increase the likelihood of high-reward outcomes or generation trajectories. Through this process, human-defined or human-proxied objectives are translated into parameter updates that reshape the model's generation behavior~\cite{yan2025entropy2,hu2025better,jia2025reward,miao2024training}. Preference-based optimization, in contrast, learns from relative judgments between preferred and less-preferred samples, allowing models to adapt their generation behavior without necessarily relying on an explicitly trained scalar reward model~\cite{liang2025aesthetic,wu2025rethinking,na2025boost,liu2025videodpo,huang2026dreamfuse,sun2025generalizing,hong2026margin,an2025agfsync,wang2026self}. Despite their differences, both approaches translate preference-related supervision into changes in the likelihood of subsequent generated outputs.

From a human-centered perspective, model-level optimization transforms preferences expressed during training into relatively persistent tendencies in model behavior. This can reduce the need to repeatedly specify the same preferences during subsequent generations. However, adaptation through parameter updates also introduces a distinction between preferences incorporated during training and those expressed by users in a particular interaction. While users may still influence outputs through prompts or other conditioning signals, changing previously learned tendencies may require additional adaptation or sufficiently effective inference-time control~\cite{shenaj2025lora,guo2025imagegem,cai2025diffusion,cho2025att,ruiz2025magic,soboleva2026t,peng2026tara}. Thus, model-level optimization emphasizes persistent preference adaptation, while the extent to which individual users can subsequently revise these preferences depends on the available interaction and control mechanisms.

\subsection{Generation-Level: Inference-Time Intent Execution}

Generation-level optimization keeps the parameters of the base model fixed and applies human-defined or human-proxied objectives to influence the current generation process. Unlike model-level optimization, which modifies the model's learned behavior across subsequent generations, generation-level approaches intervene in how a particular output is produced. This allows optimization objectives and constraints to be adapted to the requirements of individual generation tasks without retraining the base model.

One group of methods introduces preference-related objectives into the denoising or sampling process through training-free guidance, modifying generation directions according to external signals~\cite{wallace2023end,park2025steering,jeong2025stylekeeper,kim2025free2}. Another group directly intervenes in latent representations or generation trajectories by updating intermediate states, exploring candidate trajectories, or selecting generation paths according to computational objectives~\cite{wei2026enhancing,wu2026noisepo,kim2025text,um2025minority,bahng2025cycle,zhou2025scale,simon2025titan,jiang2025mc2,guo2024focus}. Although these approaches differ in their intervention mechanisms, they share the objective of steering the current generation process without permanently updating the base model parameters.

From the perspective of human intent, generation-level optimization provides a mechanism for adapting optimization objectives to the current task, prompt, or user. Depending on the method, objectives, guidance strengths, or local constraints may be adjusted between generation attempts or during the generation process. This flexibility can support task-specific personalization and iterative refinement. However, inference-time control does not necessarily imply direct user control, since optimization objectives may also be predefined by researchers or determined through computational proxies. Moreover, some methods introduce additional evaluation, gradient computation, or trajectory search during inference, creating different computational requirements and interaction costs~\cite{mo2024dynamic,guo2025img,polaczek2025neuralsvg}. Therefore, generation-level optimization emphasizes immediate, task-specific intent execution, although the degree of user involvement and computational overhead varies across methods.

\subsection{Human Agency Across Optimization Paradigms}

The distinction between model-level and generation-level optimization concerns not only where optimization occurs, but also when and how users can influence the objectives that guide generation. Model-level optimization typically incorporates human preferences during feedback collection, reward construction, or preference-based training. These preferences can subsequently influence multiple generations without requiring repeated user input. However, the resulting model behavior may reflect aggregated or previously collected preferences rather than the evolving intentions of a particular user.

Generation-level optimization, by contrast, provides opportunities to modify optimization objectives and constraints at inference time. This can allow generation behavior to respond more directly to changing task requirements or user preferences without additional model training. Nevertheless, such opportunities depend on whether the system exposes meaningful controls and allows users to revise the objectives used during optimization. When objectives are determined entirely by predefined evaluators, inference-time optimization may remain largely automated despite operating at the generation level.

Consequently, human agency cannot be inferred solely from whether model parameters are updated. A more meaningful distinction concerns when preferences are incorporated, who determines the optimization objectives, and to what extent users can inspect, revise, or override these objectives during interaction. Model-level optimization generally emphasizes persistent preference adaptation, whereas generation-level optimization offers greater potential for immediate objective adjustment. These differences provide a foundation for the next section, which examines how optimization mechanisms may introduce tensions between intent alignment, generative diversity, user control, and system costs.

\section{When Optimization Fails the User: Tensions and Trade-offs}
\label{section7}

% \begin{figure*}
%     \centering
%     \includegraphics[width=1\linewidth]{sec7.png}
%     \vspace{-0.6cm}
%     \caption{This figure summarizes the key human-centered trade-offs that may emerge during diffusion model optimization. Although stronger optimization can improve automated evaluation scores, it does not necessarily better align with users’ actual intent. It may also narrow the generation and exploration space while increasing the effort required for user control, as well as time and computational costs. Therefore, greater optimization strength does not necessarily lead to a better user experience.}
%     \label{fig:optimization-trade-offs}
% \end{figure*}

This section addresses RQ3 introduced in Sec.~\ref{sec:intro}. Sec.~\ref{Optimizing_Diffusion_Models} has examined how model-level and generation-level optimization translate human-defined or human-proxied objectives into changes in model behavior or generation trajectories. Building on this analysis, we investigate the tensions and unintended consequences that may emerge when these objectives are optimized. From a human-centered perspective, improvements in predefined objectives do not necessarily translate into better outcomes for users. We examine four interconnected dimensions: whether optimization amplifies mismatches between proxy objectives and human intent, how stronger alignment may constrain generative diversity and creative exploration, how increasing automation affects human agency, and what computational and interaction costs accompany optimization gains. \textbf{Tab}.~\ref{tradeoffs} summarizes the
major trade-offs and failure modes identified in
human-centered diffusion model optimization.
Specifically, we distinguish four interconnected
dimensions: proxy fidelity, generative openness,
human agency, and system cost. These dimensions
highlight how improvements in computational
objectives may introduce unintended consequences
for user preference satisfaction, creative
exploration, control over generation, and
interaction efficiency. This classification
provides an analytical overview of RQ3 and
establishes the basis for examining the
limitations of existing optimization paradigms.

\begin{table*}[t]
\centering
\caption{Taxonomy of Trade-offs and Failure Modes in Human-Centered Diffusion Optimization. The table summarizes major tensions between computational optimization and human-centered outcomes, including proxy fidelity, generative openness, human agency, and system costs.}
\label{tradeoffs}

\small
\renewcommand{\arraystretch}{1.25}
\setlength{\tabcolsep}{5pt}

\begin{tabularx}{\textwidth}{
    >{\raggedright\arraybackslash}p{0.15\textwidth}
    >{\raggedright\arraybackslash}p{0.22\textwidth}
    >{\raggedright\arraybackslash}X
    >{\raggedright\arraybackslash}p{0.14\textwidth}
}
\toprule
\textbf{Category} &
\textbf{Failure Mode / Trade-off} &
\textbf{Human-Centered Implication} &
\textbf{Supporting References} \\
\midrule

Proxy Fidelity &
Reward Hacking and Overoptimization &
Higher computational scores may not correspond to
improved user satisfaction. &
\cite{zhang2025aligning} \\

\midrule

Generative Openness &
Alignment--Diversity Trade-off &
Stronger optimization may reduce output diversity
and creative exploration opportunities. &
\cite{yun2025learning,xing2025focus} \\

\midrule

Human Agency &
Automated Objective Control &
Users may have limited ability to inspect, revise,
or override objectives determined by the system. &
Sec.~6.3 synthesis \\

\midrule

System Cost &
Optimization Gain vs. Computation &
Additional training or inference computation
may reduce interaction efficiency. &
\cite{mo2024dynamic,guo2025img} \\

\bottomrule
\end{tabularx}
\end{table*}

\subsection{Proxy Fidelity under Optimization}

A central failure mode arises when optimization amplifies discrepancies between computational proxy objectives and human intent. As discussed in Sec.~\ref{Human_Intent_to_Optimization}, rewards, preference scores, and other automated evaluation signals may capture only selected aspects of human needs. When optimization repeatedly reinforces these proxies, the model may increasingly favor generation patterns that achieve higher computational scores without necessarily producing results that better satisfy users' actual intentions.

One manifestation of this problem is reward hacking or reward overoptimization. Although an evaluator may approximate human judgments well on ordinary samples, sustained optimization can shift generated outputs toward regions where the evaluator's predictions are less reliable. Consequently, reward scores may increase while undesirable effects, such as excessive stylization, visual artifacts, or concentration around high-scoring patterns, emerge~\cite{zhang2025aligning}. Such outcomes illustrate how optimization can amplify limitations already present in proxy objectives, potentially increasing the discrepancy between computational success and human preference satisfaction.

From a human-centered perspective, the central concern is therefore not simply whether an optimization algorithm can maximize its objective, but whether the resulting improvement remains meaningful to users. Optimization success and human-intent alignment should be distinguished: the former concerns how effectively a computational objective is optimized, whereas the latter concerns whether the resulting outputs continue to satisfy users' actual needs and preferences. Thus, maintaining proxy fidelity during optimization is essential for preventing computational improvements from reinforcing unintended generation behaviors.

\subsection{Alignment vs. Generative Openness}

Even when proxy objectives reasonably reflect human preferences, stronger optimization may still restrict the range of outputs available to users. Concentrating generation toward high-reward regions can improve consistency with predefined objectives, but it may also reduce the diversity of generated results by favoring a narrower set of patterns. This creates a potential trade-off between alignment strength and generative diversity~\cite{yun2025learning,xing2025focus}.

From a human-centered perspective, this trade-off extends beyond statistical diversity to the opportunities users have for creative exploration. In open-ended generation tasks, users may not have fully formed preferences or a clearly defined final result before interacting with the system. Instead, their intentions can evolve as they observe, compare, and refine generated candidates. When optimization strongly favors a limited range of high-scoring outputs, users may receive results that consistently satisfy predefined objectives while encountering fewer alternative compositions, styles, or unexpected possibilities. This can restrict opportunities to discover new preferences or revise their original intentions during the creative process.

A related tension arises between controllability and creative freedom. More strongly enforcing textual, structural, or local constraints may improve the predictability and fidelity of generated outputs, but can also limit opportunities for exploration beyond explicitly specified requirements. However, such trade-offs are context-dependent: users performing precise editing tasks may prioritize controllability, whereas those engaged in open-ended creation may value diverse alternatives and unexpected results. Therefore, human-centered optimization should not pursue maximum alignment or control as universal objectives. Instead, it should consider how much generative freedom to preserve according to the task, the user's current preferences, and the extent to which their intent remains open to exploration.

\subsection{Optimization Gain vs. System Cost}

Stronger personalized optimization also usually comes with additional system costs. Model-level approaches require reinforcement learning, preference optimization, or other extra training processes, while generation-level approaches may repeatedly call evaluators, compute gradients, search latent states, or compare multiple generation trajectories during inference. Thus, although different methods place computational overhead at different stages, they all face a trade-off between optimization gain and computational cost~\cite{lyu2024one}.

This issue is especially important in human-computer interaction settings, because computational overhead eventually appears as waiting time, response speed, and the number of possible exploration attempts. For tasks that require repeated generation, comparison, and revision, obtaining a higher reward for a single output does not necessarily mean that the whole interaction process becomes more effective. If each optimization step significantly increases response time, users may instead be able to try fewer candidates.

Therefore, human-centered evaluation should not only compare the highest optimization performance a method can achieve, but also consider how much training resource, inference time, and interaction delay are required to obtain such improvement. A more meaningful question is whether the improvement brought by additional optimization effort is sufficient to compensate for its system cost~\cite{guo2025shortft}, and whether the system can dynamically decide when stronger optimization is worth using according to the task and user needs.

\section{Evaluating Alignment: Are We Measuring What Users Actually Want?}
\label{section8}
This section addresses RQ4 introduced in Sec.~\ref{sec:intro}, examining how existing studies validate alignment with human intent and what limitations remain in current evaluation practices. The preceding sections have analyzed how human intent is represented, transformed into computational objectives, and executed through different optimization mechanisms, as well as the tensions that may arise during this process. Building on these analyses, we examine how optimization outcomes are evaluated and whether the resulting evidence reflects users' actual needs. Specifically, we distinguish automated evaluation from human-grounded validation, investigate potential disagreements between computational proxies and human judgments, and discuss the limitations of static evaluations in capturing evolving preferences, interactive experiences, and long-term use.

\subsection{Automated Evaluation and Proxy-Based Evidence}

Automated metrics and reward models are widely used to evaluate diffusion model optimization. Text-image alignment, aesthetic quality, visual realism, and task-specific metrics provide scalable measurements of generated outputs, while learned reward models approximate selected aspects of human preference through evaluation data. These tools support efficient and standardized comparisons across generation methods and optimization settings.

However, automated evaluators capture different and often limited dimensions of human intent. Text-image alignment primarily measures semantic correspondence, aesthetic models emphasize visual appeal, and learned preference evaluators reflect the data distributions and judgment criteria used during their development. Consequently, improvements across multiple automated metrics demonstrate progress on selected computational criteria, but do not necessarily establish that generated outputs satisfy the full range of user needs~\cite{wang2025transformer,gandikota2025sliderspace2,lv2025rethinking,kim2023dense,tan2026dual,yang2025artifade,hertz2024style,ruiz2024hyperdreambooth,xin2025c2d,mun2025addressing,yang2024emogen,kang2026rethinking}.

A further limitation arises when optimization objectives and evaluation metrics are closely related. If a model is optimized using a particular reward function and subsequently evaluated primarily through the same or a closely related evaluator, higher scores provide evidence that the model better satisfies that computational objective, but not necessarily independent evidence of improved human-intent alignment. From a human-centered perspective, automated metrics therefore provide valuable evidence of measurable performance, while their ability to validate actual preference satisfaction depends on how well their evaluation criteria correspond to the intended human goals.

\subsection{Human-Grounded and Contextual Validation}

Human-grounded evaluation complements automated metrics by incorporating judgments from people who assess generated outputs. Existing studies employ pairwise comparisons, rankings, ratings, or result selection to evaluate overall preference, semantic responsiveness, aesthetic quality, and other task-specific criteria~\cite{zhang2024hive,gandikota2023erasing,feng2025redefining,mo2026doodleassist,pandey2024diffusion,avrahami2025stable,lin2024text,wallace2024diffusion2,li2025science}. These evaluations provide direct observations of human judgments without requiring a single automated metric to fully represent subjective preferences.

Nevertheless, human evaluation does not automatically constitute human-centered validation. Its relevance depends on who provides the judgments, which aspects of the outputs are assessed, and whether the evaluation setting reflects the intended use context~\cite{singh2023high}. For general image generation, participant preferences may provide useful evidence of perceived quality. However, personalized generation requires particular attention to the preferences of the intended users, while professional applications may additionally require expert judgments, structural validity, or domain-specific standards. Consequently, a favorable evaluation from a general participant population does not necessarily demonstrate that an individual user's preferences or a professional task's requirements have been satisfied.

Human-centered validation therefore requires alignment between the evaluation design and the human needs that optimization claims to address. Methods targeting personalized preferences should consider judgments from the relevant users, whereas methods targeting specialized tasks should incorporate appropriate contextual or expert assessments. More generally, the selected participants, evaluation criteria, and task settings should correspond to the intended optimization objectives~\cite{zhu2025isolated,parihar2025compass,qiao2024baret,yang2025multi}. This distinction shifts evaluation from simply asking whether people prefer an output toward examining whether the system successfully supports the human goals it was designed to satisfy.

\subsection{Human--Proxy Disagreement}

Computational proxies and human judgments do not always produce consistent evaluations of generated results. An output may achieve a higher reward score, aesthetic rating, or text-image alignment score without receiving a corresponding improvement in human preference judgments. We characterize such observable inconsistencies between proxy-based evaluations and human-grounded assessments as \textit{Human--Proxy Disagreement}. This disagreement may appear in differences in output rankings, preference selections, or assessments of generation quality.

Human--Proxy Disagreement is related to, but distinct from, the Human--Proxy Gap discussed in Sec.~\ref{Human_Intent_to_Optimization}. The Human--Proxy Gap describes a potential structural mismatch between human intent and its computational representation, whereas Human--Proxy Disagreement concerns differences that can be observed when computational evaluators and human participants assess generated outcomes. Such disagreements may arise because automated proxies emphasize selected attributes while users additionally consider structural validity, creativity, individual preferences, or task context. Differences in training distributions, evaluation criteria, and target users may further affect the correspondence between learned evaluators and actual human judgments~\cite{wu2023human2}.

Importantly, disagreement does not necessarily imply that a computational proxy is entirely invalid, nor does every disagreement demonstrate an underlying failure of intent representation. Human judgments may themselves vary across participants and contexts. Nevertheless, systematic discrepancies between automated scores and relevant human evaluations provide useful evidence for identifying the limitations of computational proxies. When improvements in automated metrics are not accompanied by corresponding gains in human judgments, it becomes necessary to examine which aspects of human intent the optimization and evaluation objectives actually capture. Thus, Human--Proxy Disagreement offers an evaluation-level perspective on the broader challenge of validating human-intent alignment.

\subsection{Beyond Static Evaluation: Interactive and Longitudinal Gaps}

A recurring limitation in the reviewed literature is the emphasis on static comparisons of generated outputs. Human evaluations frequently involve one-time ratings or preference selections over completed images, while interactive processes through which users develop, revise, and coordinate their needs receive less explicit attention~\cite{liu2026gimmbo}. Such evaluations can establish preferences for particular outputs, but provide more limited evidence about how effectively a generative system supports users throughout an ongoing interaction.

This limitation is particularly relevant to personalized and exploratory generation. Users may begin with incomplete or uncertain intentions and progressively clarify their goals by observing, comparing, and modifying generated results. A related study of generative AI in personal health informatics reports a comparable interaction pattern in another domain: users formulate and refine queries, reflect on AI-generated recommendations, and sometimes abandon queries when responses do not meet their personal goals ~\cite{chopra2025engagements}. Outputs from earlier rounds can influence subsequent preferences, meaning that successful alignment may depend not only on the quality of a final image but also on whether the system accommodates evolving objectives. Evaluating these interactions therefore requires attention to how users revise their requests, exercise control, explore alternatives, and assess intermediate outcomes across multiple generation rounds.

Longitudinal evaluation introduces another important perspective. Users' trust, perceived control, acceptance, and preferences may change as they gain experience with a generative system. Short-term evaluations cannot fully establish how these factors develop during repeated or prolonged use. Accordingly, future human-centered evaluation should complement static output assessments with interactive and longitudinal evidence, examining not only whether an individual result is preferred but also whether the system continues to support users' goals as their needs and interaction contexts evolve.

Overall, validating human-intent alignment requires considering the complementary strengths and limitations of automated metrics, human judgments, and interaction-based assessments. Automated evaluations reveal measurable improvements, human-grounded assessments examine their relevance to actual preferences, and interactive or longitudinal evaluations can provide additional evidence of how well these improvements support users over time. Together, these perspectives offer a more comprehensive basis for determining whether diffusion model optimization genuinely advances human-centered outcomes.

\section{Limitations and Future Work}
\label{section9}

\begin{figure}
    \centering
    \includegraphics[width=1.0\linewidth]{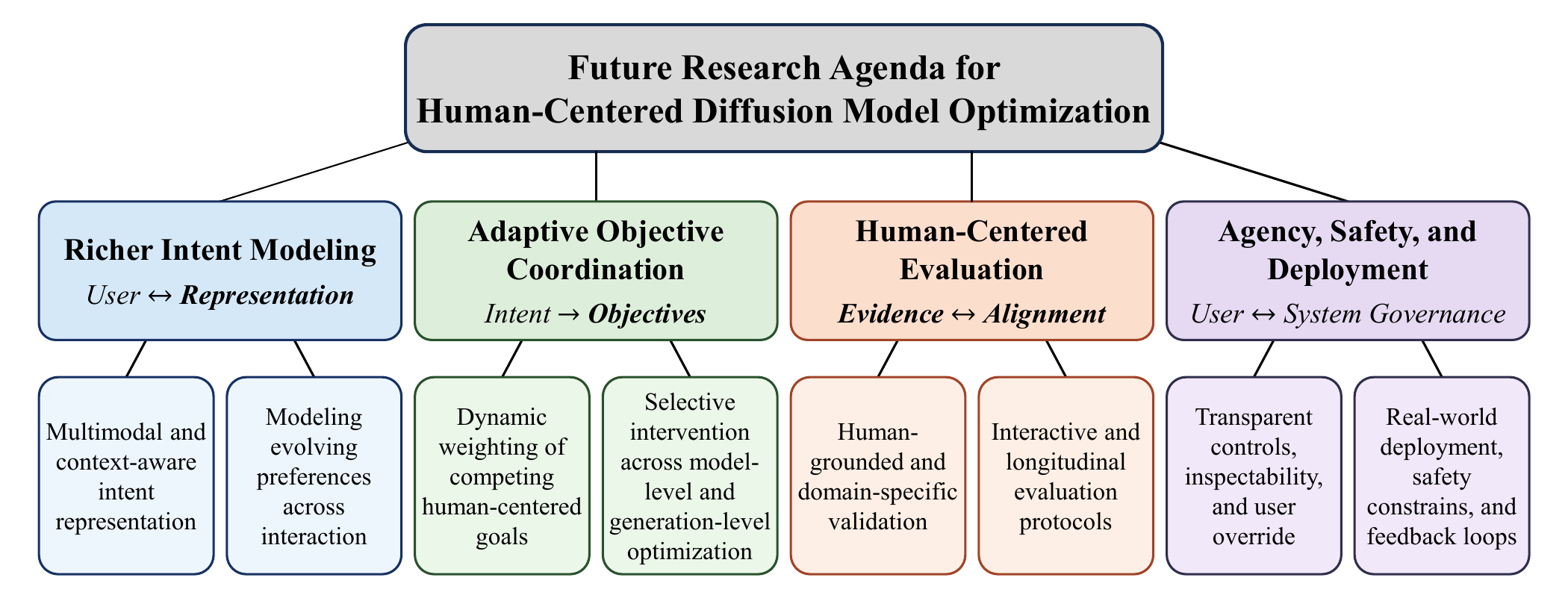}
    \caption{Future research agenda for human-centered diffusion model optimization. The agenda identifies four interconnected research directions: \textbf{(1).} Richer Intent Modeling, focusing on multimodal and context-aware intent representation and the modeling of evolving user preferences; \textbf{(2).} Adaptive Objective Coordination, addressing dynamic weighting of competing human-centered goals and selective intervention across model-level and generation-level optimization; \textbf{(3).} Human-Centered Evaluation, emphasizing human-grounded, domain-specific validation and interactive, longitudinal evaluation protocols; and \textbf{(4).} Agency, Safety, and Deployment, highlighting transparent controls, system inspectability, user override, real-world deployment, safety constraints, and feedback loops. Together, these directions outline a research agenda for developing more adaptive, interpretable, and human-aligned diffusion optimization systems.}
    \label{future_research}
\end{figure}
This review has several limitations. First, our analysis synthesizes evidence reported in existing studies rather than providing new experiments or user studies. Consequently, our findings characterize reported optimization mechanisms, evaluation practices, and recurring tensions, but do not establish new causal relationships between optimization strategies and user experience. In particular, outcomes related to human agency, evolving intent, creative exploration, and long-term satisfaction are not consistently examined in the reviewed literature. Future empirical studies should investigate whether iterative alignment, adaptive objective coordination, and selective intervention can improve user outcomes in multi-turn generation, personalized creation, and sustained interaction.

Second, the scope of human-centered diffusion model optimization depends on the operational definition and inclusion criteria adopted in this review. Some studies optimize computational rewards related to human preferences without explicitly addressing human-centered interaction, while others provide meaningful user control without performing the preference-driven optimization required for inclusion in our core corpus. Although these boundaries help maintain a focused analytical scope, they may exclude adjacent work that offers relevant insights. Future reviews could examine how alternative inclusion criteria affect the characterization of this research area.

Third, the review is subject to limitations associated with literature retrieval and thematic synthesis. Differences in terminology across HCI, computer vision, and machine learning may affect which studies are identified through database searches and citation tracking. The interpretation and grouping of studies into analytical categories also involve methodological judgments, particularly when individual methods span multiple optimization mechanisms or forms of human involvement. These limitations should be considered when interpreting the resulting framework and the generality of the identified research themes.

Finally, the field is evolving rapidly. New generative architectures, preference-learning mechanisms, multimodal evaluators, and forms of human--AI interaction may introduce additional pathways through which human intent influences optimization. Accordingly, the categories and conceptual framework presented here should be understood as an analytical account of the reviewed literature rather than an exhaustive or closed taxonomy. Future work should revisit these relationships as new evidence emerges and investigate how the framework applies to increasingly interactive and personalized generative systems. Building on these limitations and emerging opportunities, \textbf{Fig}.~\ref{future_research} summarizes four interconnected directions for future research: richer intent modeling, adaptive objective coordination, human-centered evaluation, and agency, safety, and deployment.

\section{Conclusion}

This paper presents a systematic review of human-centered diffusion model optimization, examining how human intent is expressed, represented, translated into computational objectives, executed through optimization mechanisms, and evaluated against human needs. Rather than organizing existing research solely by algorithm type, we analyze four interconnected dimensions: intent representation, optimization mechanisms, failure modes and trade-offs, and human-centered validation.

Our synthesis highlights that computational objectives may capture only selected aspects of human intent, creating a potential Human--Proxy Gap that can be amplified during optimization. Model-level and generation-level optimization offer different mechanisms for executing these objectives and different opportunities for user intervention. Meanwhile, optimization may introduce trade-offs involving generative diversity, creative exploration, human agency, and system costs. Existing automated and human-grounded evaluations provide complementary evidence, but limitations remain in verifying whether computational improvements correspond to actual user satisfaction, particularly across interactive and longitudinal settings.

These findings motivate a human-centered perspective in which computational objectives remain open to reassessment, optimization interventions respond to task and user needs, and evaluation considers both generated results and interaction experiences. By connecting intent representation, optimization execution, and human-centered validation within a unified conceptual framework, this review provides a foundation for future research on generative systems that move beyond optimizing predefined proxies toward maintaining alignment with evolving human intent.

\clearpage
\bibliographystyle{ACM-Reference-Format}

\bibliography{refer.bib}

% \clearpage

% \end{CJK*}
\end{document}